\documentclass[runningheads]{llncs}
\usepackage[mobile]{eccv}
\usepackage{eccvabbrv}
\usepackage{graphicx}
\usepackage{booktabs}
\usepackage[accsupp]{axessibility}  %

\usepackage{eccvabbrv}
\usepackage{graphicx}
\usepackage{booktabs}

\definecolor{linkcolor}{HTML}{ED1C24}
\usepackage[pagebackref,breaklinks,colorlinks,citecolor=eccvblue]{hyperref}
\usepackage{orcidlink}
\usepackage{url}
\usepackage{graphicx}
\usepackage{wrapfig}
\usepackage{tabularx}
\usepackage{color, colortbl}
\usepackage{physics}
\usepackage{booktabs}
\usepackage{multirow}
\usepackage{xcolor}
\usepackage{graphicx}
\usepackage{footnote}
\usepackage{tabularx}
\usepackage{float}
\usepackage{lipsum}
\usepackage{multicol}
\usepackage{siunitx}
\usepackage{tabularx,booktabs}
\usepackage{etoolbox}
\usepackage{booktabs}
\usepackage{threeparttable}

\usepackage{amsmath, amssymb, caption, subcaption, multirow, overpic, textpos}

\newcommand{\ours}{DVT\xspace}
\newcommand{\ourslong}{Denoising Vision Transformers\xspace}

\newlength\savewidth\newcommand\shline{\noalign{\global\savewidth\arrayrulewidth
  \global\arrayrulewidth 1pt}\hline\noalign{\global\arrayrulewidth\savewidth}}
\newcommand{\tablestyle}[2]{\ttfamily\setlength{\tabcolsep}{#1}\renewcommand{\arraystretch}{#2}\centering\scriptsize}
\definecolor{baselinecolor}{gray}{.9}
\newcommand{\baseline}[1]{\cellcolor{baselinecolor}{#1}}

\newcolumntype{x}[1]{>{\centering\arraybackslash}p{#1pt}}
\newcolumntype{y}[1]{>{\raggedright\arraybackslash}p{#1pt}}
\newcolumntype{z}[1]{>{\raggedleft\arraybackslash}p{#1pt}}

\usepackage[pagebackref,breaklinks,colorlinks,citecolor=eccvblue]{hyperref}
\usepackage{hyperref}

\usepackage{orcidlink}
\usepackage{python}

\definecolor{darkpastelgreen}{rgb}{0.01, 0.75, 0.24}
\definecolor{darkgray}{rgb}{0.66, 0.66, 0.66}
\definecolor{bittersweet}{rgb}{1.0, 0.44, 0.37}
\definecolor{bleudefrance}{rgb}{0.19, 0.55, 0.91}
\definecolor{ao(english)}{rgb}{0.0, 0.5, 0.0}

\newcolumntype{x}[1]{>{\centering\arraybackslash}p{#1pt}}
\newcolumntype{y}[1]{>{\raggedright\arraybackslash}p{#1pt}}
\newcolumntype{z}[1]{>{\raggedleft\arraybackslash}p{#1pt}}
\usepackage[capitalize]{cleveref}
\newcommand{\app}{\raise.17ex\hbox{$\scriptstyle\sim$}}

\definecolor{deemph}{gray}{0.7}
\newcommand{\gc}[1]{\textcolor{gray}{#1}}

\newcommand{\increase}[1]{{\color{darkpastelgreen} #1}}
\newcommand{\decrease}[1]{{\color{darkgray} #1}}

\newcolumntype{x}[1]{>{\centering\arraybackslash}p{#1pt}}
\newcolumntype{y}[1]{>{\raggedright\arraybackslash}p{#1pt}}
\newcolumntype{z}[1]{>{\raggedleft\arraybackslash}p{#1pt}}

{
\setlength{\leftmargini}{9pt}%
\begin{itemize}%
\setlength{\itemsep}{0pt}%
\setlength{\topsep}{0pt}%
\setlength{\partopsep}{0pt}%
\setlength{\parskip}{0pt}}%
{\end{itemize}}

\renewcommand{\paragraph}[1]{\vspace{1.25mm}\noindent\textbf{#1}}

\graphicspath{{./figures/}}

\definecolor{brandeisblue}{rgb}{0.0, 0.44, 1.0}
\definecolor{burntorange}{rgb}{0.8, 0.33, 0.0}

\begin{document}

\title{\ourslong}
\titlerunning{\ourslong}

\author{
    Jiawei Yang$^{*,\dagger,1,}$\thanks{Work partially completed while interning at Google Research} \hspace{2.5mm}
    Katie Z Luo$^{*,2}$ \hspace{2.5mm}
    Jiefeng Li$^3$ \hspace{2.5mm}
    Congyue Deng$^4$ \\
    Leonidas Guibas$^4$ \hspace{2.5mm}
    Dilip Krishnan$^5$ \hspace{2.5mm}
    Kilian Q Weinberger$^2$\\
    Yonglong Tian$^5$ \hspace{2.5mm}
    Yue Wang$^1$
}
\institute{
    $^1$University of Southern California \hspace{2.5mm} $^2$Cornell University
    \\ $^3$Shanghai Jiaotong University \hspace{2.5mm} $^4$Stanford University
    \\
    $^5$Google DeepMind \\
    $^*$equal technical contribution \hspace{5mm} $^\dagger$project lead
}
\authorrunning{J.\ Yang, K.\ Luo et al.}

\maketitle

\begin{center}
    \captionsetup{type=figure}
    \includegraphics[width=0.99\linewidth]{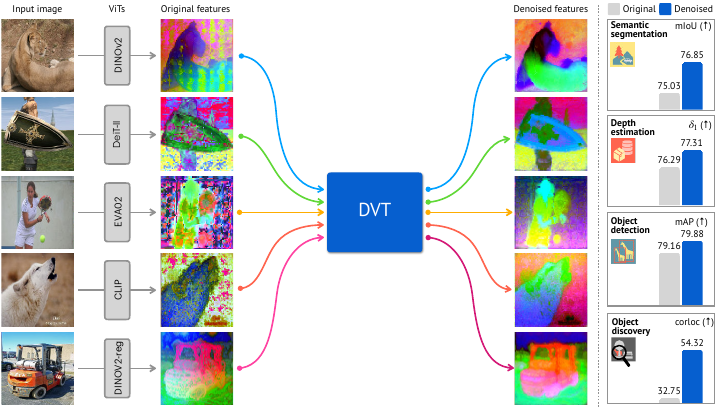}
    \captionof{figure}{\textbf{\ourslong (\ours)} effectively suppresses noisy artifacts in the visual features of all Vision Transformers (ViTs) we have tested and improves performance on a broad spectrum of dense prediction tasks, including semantic segmentation, depth estimation, object detection, and object discovery.
    Our evaluation encompasses a representative set of ViTs, including DINOv2 \cite{oquab2023dinov2}, DeiT-III \cite{touvron2022deit}, EVA-02 \cite{fang2023eva}, CLIP \cite{radford2021learning}, and DINOv2-reg \cite{darcet2023vision}.
        We visualize the features before and after \ours{}, colored via principal component analysis (PCA). Best viewed in color. \textbf{Right}: We report the downstream dense prediction task performances, averaged over all models.}
    \label{fig:teaser}
\end{center}

\begin{abstract}
    We study a crucial yet often overlooked issue inherent to Vision Transformers (ViTs): feature maps of these models exhibit grid-like artifacts (``Original features'' in \cref{fig:teaser}), which hurt the performance of ViTs in downstream dense prediction tasks such as semantic segmentation, depth prediction, and object discovery.
    We trace this issue down to the positional embeddings at the input stage.
    To mitigate this, we propose a two-stage denoising approach, termed \ourslong (\ours).
    In the first stage, we separate the clean features from those contaminated by positional artifacts by enforcing cross-view feature consistency with neural fields on a per-image basis. This per-image optimization process extracts artifact-free features from raw ViT outputs, providing clean feature estimates for offline applications. In the second stage, we train a lightweight transformer block to predict clean features from raw ViT outputs, leveraging the derived estimates of the clean features as supervision. Our method, \ours{}, does not require re-training the existing pre-trained ViTs, and is immediately applicable to any Vision Transformer architecture. We evaluate our method on a variety of representative ViTs (DINO, DeiT-III, EVA02, CLIP, DINOv2, DINOv2-reg) and demonstrate that \ours consistently improves existing state-of-the-art general-purpose models in semantic and geometric tasks across multiple datasets (\cref{fig:teaser}, right, \cref{tab:dense_results,tab:obj_det,tab:obj_discovery}).
    We hope our study will encourage a re-evaluation of ViT design, especially regarding the naive use of positional embeddings. Our code and checkpoints are publicly available in our \href{https://jiawei-yang.github.io/DenoisingViT/}{project page}.
\end{abstract}
\section{Introduction}
\label{sec:intro}

In recent years, Transformers \cite{vaswani2017attention} have emerged as the universal architecture for modern foundation models across many modalities, from text \cite{raffel2023exploring, chowdhery2022palm, radford2018improving,brown2020language} to audio \cite{li2018close, Wang_2017}, and images \cite{dosovitskiy2021image, Carion_2020}. Among these, Vision Transformers (ViTs) \cite{dosovitskiy2021image}
trained at scale not only achieve state-of-the-art under multiple benchmarks but also exhibit intriguing behaviors and capabilities across various tasks \cite{caron2021emerging,he2022masked,radford2021learning,oquab2023dinov2}.

\begin{figure}[t!]
    \centering
    \includegraphics[width=\linewidth]{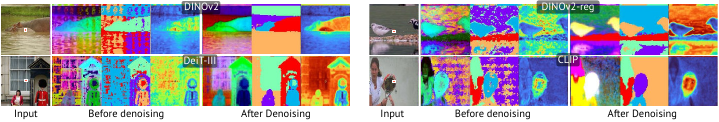}
    \caption{\textbf{Artifacts hurt semantic coherence.} For each triplet, we show a feature map, a K-Means cluster map, and a similarity map of the central patch (red dotted) with other patches in the image. Observe how artifacts negatively impact clustering accuracy and similarity correspondences, and how our denoising mitigates these issues.}
    \label{fig:clustering}
\end{figure}

Despite these significant strides made by ViTs, our work reveals a crucial yet often overlooked challenge: the presence of persistent noise artifacts in ViT outputs, observable across various training algorithms \cite{dosovitskiy2021image, oquab2023dinov2,touvron2022deit,radford2021learning,fang2023eva,caron2021emerging} (illustrated in \cref{fig:teaser} left). These artifacts not only compromise visual clarity but also hinder feature interpretability and disrupt semantic coherence.
For example, \cref{fig:clustering} demonstrates that applying clustering algorithms directly on the raw ViT output results in noisy clusters, and the patch feature similarity is less reliable. 
Additionally, these artifacts are frequently concealed by \textit{seemingly} impressive performance on downstream tasks, thus evading thorough examination or detection by the research community.
Addressing these artifacts can unleash the potential of pre-trained ViTs and lead to substantial performance improvements (\cref{fig:teaser} right). Therefore, our work aims to answer a crucial research question: \textit{Is it feasible to effectively denoise these artifacts in pre-trained ViTs, ideally without model retraining?}

\begin{figure}[t!]
    \centering
    \includegraphics[width=\linewidth]{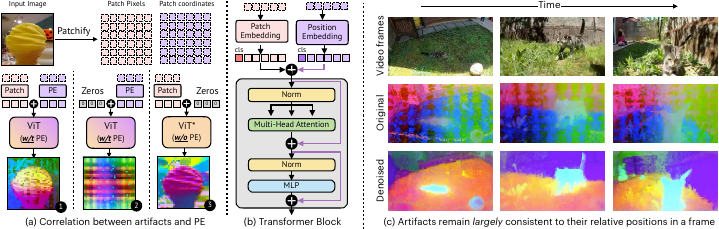}
    \caption{
        \textbf{Impact of positional embeddings in ViTs.} (a) Comparison between DINOv2 ViTs \cite{oquab2023dinov2} trained with and without positional embeddings ((``ViT'' v.s. ``ViT$^*$''). We show feature maps from (1) a standard ViT, (2) a ViT using only positional embeddings (PE) as input, emphasizing the emergence of artifacts, and (3) a PE-free ViT$^*$, displaying a clear absence of these artifacts. In the figure, ``Patch'': patch embedding, ``PE'': position embedding. (b) Illustration of how ViT retains and propagates the positional embeddings. (c) Despite significant differences in the context of various frames, the artifacts largely maintain a consistent relative position in the images (central row). Our \ours effectively denoises these artifacts, demonstrated in the final row.
    }
    \label{fig:vit_arch}
\end{figure}

To answer this, we first investigate the origins of these artifacts. We hypothesize that positional embeddings, a fundamental component of ViT architecture, play a pivotal role in the emergence of these artifacts. Our initial analysis supports this hypothesis: \textit{First}, when a zero tensor is fed into a pre-trained DINOv2 model \cite{oquab2023dinov2}, the resulting output is predominantly characterized by similar noise patterns (\cref{fig:vit_arch}-(a, 2)). \textit{Second}, we observe the absence of such artifacts in the outputs of a DINOv2 model trained without positional embeddings,  which contrasts sharply with the standard model outputs (\cref{fig:vit_arch}-(a, 1) v.s. (a, 3)).
\textit{Third}, take a video with continuous frames as an example (\cref{fig:vit_arch}-(c)). Despite the significant differences in the context of various input frames, the artifacts maintain a generally consistent relative position in the images (\cref{fig:vit_arch}-(c), middle row).

With these insights, we present a two-stage denoising approach, \ourslong (\ours), to suppress artifacts in pre-trained ViTs. In the first stage, we obtain clean features from contaminated ones by enforcing cross-view feature consistency and artifact consistency with neural fields on a per-image basis. This per-image denoising process extracts noise-free features from raw output, providing these denoised ViT features for offline applications. In the second stage, we train a lightweight denoiser model, consisting of a single transformer block, to predict the denoised features from the raw ViT outputs. More importantly, this denoiser can be seamlessly integrated into pre-trained ViTs without extensive \textit{re-training}, providing denoised features for online applications and generalizing well to unseen data.

We conduct empirical evaluations to demonstrate the efficacy of \ours on six representative ViTs: DINO \cite{caron2021emerging}, DINOv2 \cite{oquab2023dinov2}, DINOv2 with Register (DINOv2-reg) \cite{darcet2023vision}, DeiT-III \cite{touvron2022deit}, EVA-02 \cite{fang2022eva,fang2023eva}, and CLIP \cite{radford2021learning}. These evaluations demonstrate significant improvements in performance across various dense prediction vision tasks such as semantic segmentation, depth estimation, object detection, and object discovery. In summary, our contributions are:
\begin{itemize}
    \item We identify and highlight the widespread occurrence of noise artifacts in ViT features, pinpointing positional embeddings as a crucial underlying factor. To the best of our knowledge, we are the first to provide such an analysis.
    \item We introduce a tailored noise model for ViTs, along with a neural field based denoising technique. This combination effectively isolates and removes noise artifacts from ViT features.
    \item We develop a flexible and efficient denoiser that integrates seamlessly with pre-trained ViTs, enabling real-time applications.
    \item Our approach results in substantial performance improvements across various ViTs and downstream dense prediction tasks (\cref{fig:teaser}, right, \cref{tab:dense_results,tab:obj_det,tab:obj_discovery}).
\end{itemize}

\section{Related Works}
\label{sec:related}

\paragraph{General purpose features from Vision Transformers.} Transformers have been used extensively across multiple domains as general-purpose feature extractors \cite{brown2020language,radford2019language,chowdhery2022palm,touvron2023llama,devlin2019bert, raffel2023exploring}.
Vision Transformers \cite{dosovitskiy2021image} (ViTs) pre-trained via supervised learning \cite{vaswani2023attention, touvron2022deit,kirillov2023segment} or self-supervised learning \cite{zhou2021ibot,he2022masked,caron2021emerging,oquab2023dinov2} have demonstrated strong generalizability to various downstream visual tasks, even without fine-tuning. However, we show that ViTs trained with diverse training objectives exhibit commonly observed noise artifacts in their output feature maps. These artifacts are often overlooked in practice because their presence cannot be simply reflected by image \textit{classification} accuracy. Thus, our work focuses on evaluating pre-trained ViTs for \textit{dense recognition} tasks such as segmentation, depth estimation, and object discovery. We demonstrate how these artifacts adversely affect dense recognition tasks, thereby motivating our method to mitigate them.

\paragraph{ViT artifacts.} Our work studies the noise artifacts in ViTs, an issue that has been previously observed but often remains unexplored. These artifacts manifest as noisy attention maps in supervised ViTs (\ie, ViTs do not attend to objects of interest well) \cite{caron2021emerging,chen2021vision}. Concurrently with our study, two recent studies similarly have also identified artifacts in self-supervised ViTs \cite{yang2023emernerf,darcet2023vision}. Specifically, \cite{darcet2023vision} describe these as ``high-norm'' patches in low-informative background regions, hypothesizing their occurrence is limited to large (\eg ViT-\textit{large} or greater) and sufficiently trained ViTs. However, our analysis indicates that this may not be the \textit{full picture}, as we observe similar artifacts in small or base ViTs that cannot be easily identified by extremely high feature norm values. Instead, we find a strong correlation between the presence of artifacts and the use of positional embeddings in ViTs. This finding suggests that artifacts are not strictly confined to certain model sizes or training scales but are more fundamentally linked to the inherent design of ViTs.
Moreover, unlike the method proposed by \cite{darcet2023vision} that retrains ViTs with register tokens \cite{goyal2023think, xiao2023efficient} from scratch,
our approach directly denoises pre-trained models without retraining. Users can \textit{dynamically} enable or disable the plugged-in denoiser as needed. Lastly, we note that some \textit{weak} artifacts still exist in DINOv2 models trained with registers \cite{darcet2023vision} (see \cref{fig:teaser} DINOv2-reg and appendix), and our \ours can effectively denoise them, improving the performance of DINOv2-reg.

\section{Preliminaries}
\label{sec:background}

\paragraph{Forward process in ViTs.} Despite varying training approaches, the ViT architecture has mostly remained consistent with its original design as presented in \cite{dosovitskiy2021image} and \cite{vaswani2023attention}. The forward process of a ViT, depicted in \cref{fig:vit_arch}-(b), starts by converting images into 2D patches and then embedding them, followed by a forward process of Transformer blocks. Specifically, an image $\vb{x} \in \mathbb{R}^{H\times W \times C}$ is first divided into patches $\vb{x}_{p} \in \mathbb{R}^{N\times (P^2 \cdot C)}$, where $(H, W)$ denotes the image resolution, $P$ is the patch resolution, $C$ represents the number of pixel channels and $N$ is the number of total patches. These patches are then mapped to $D$ dimensions using a trainable linear projection $\vb{E} \in {\mathbb{R}^{(P^2 \cdot C)\times D}}$ to generate patch embeddings. To inject spatial information, positional embeddings, which encode patch coordinates and are denoted $\vb{E}_{pos}^{i}$, are added to the patch embeddings. Formally, the forward process of a ViT is as follows:
\begin{align}
    \vb{z}_0  & = [ \vb{x}_{\text{cls}} + \vb{E}_{pos}^{\text{cls}} ; \vb{x}_p^0\vb{E} + \vb{E}_{pos}^{0} ; ~\cdots ; ~\vb{x}_p^{N-1}\vb{E} + \vb{E}_{pos}^{N-1}]  \label{eq1} \\
    \vb{z'}_l & =\text{MSA}\left( \text{LN} (\vb{z}_{l-1}) \right) + \vb{z}_{l-1}, \quad l= 1 \cdots L \label{eq2}                                                             \\
    \vb{z}_l  & =\text{MLP}\left( \text{LN} (\vb{z'}_{l}) \right) + \vb{z'}_{l}, \quad\quad ~~l= 1 \cdots L \label{eq3}                                                        \\
    \vb{y}    & = \text{LN} (\vb{z}_{L}) \label{eq4}
\end{align}
Here, $\vb{x}_{\text{cls}}$ and $\vb{E}_{pos}^{\text{cls}}$ represent the class token and its positional embedding, respectively, $L$ denotes the number of layers, and LN stands for layer normalization. Multi-head self-attention layers and multi-layer perceptron layers are termed MSA and MLP, respectively. Note how the \textit{input-independent} positional embeddings function as a spatial inductive basis, intermixing with inputs and propagating throughout ViT.

\section{\ourslong}
\label{sec:method}
\begin{figure}[t!]
    \centering
    \includegraphics[width=1\linewidth]{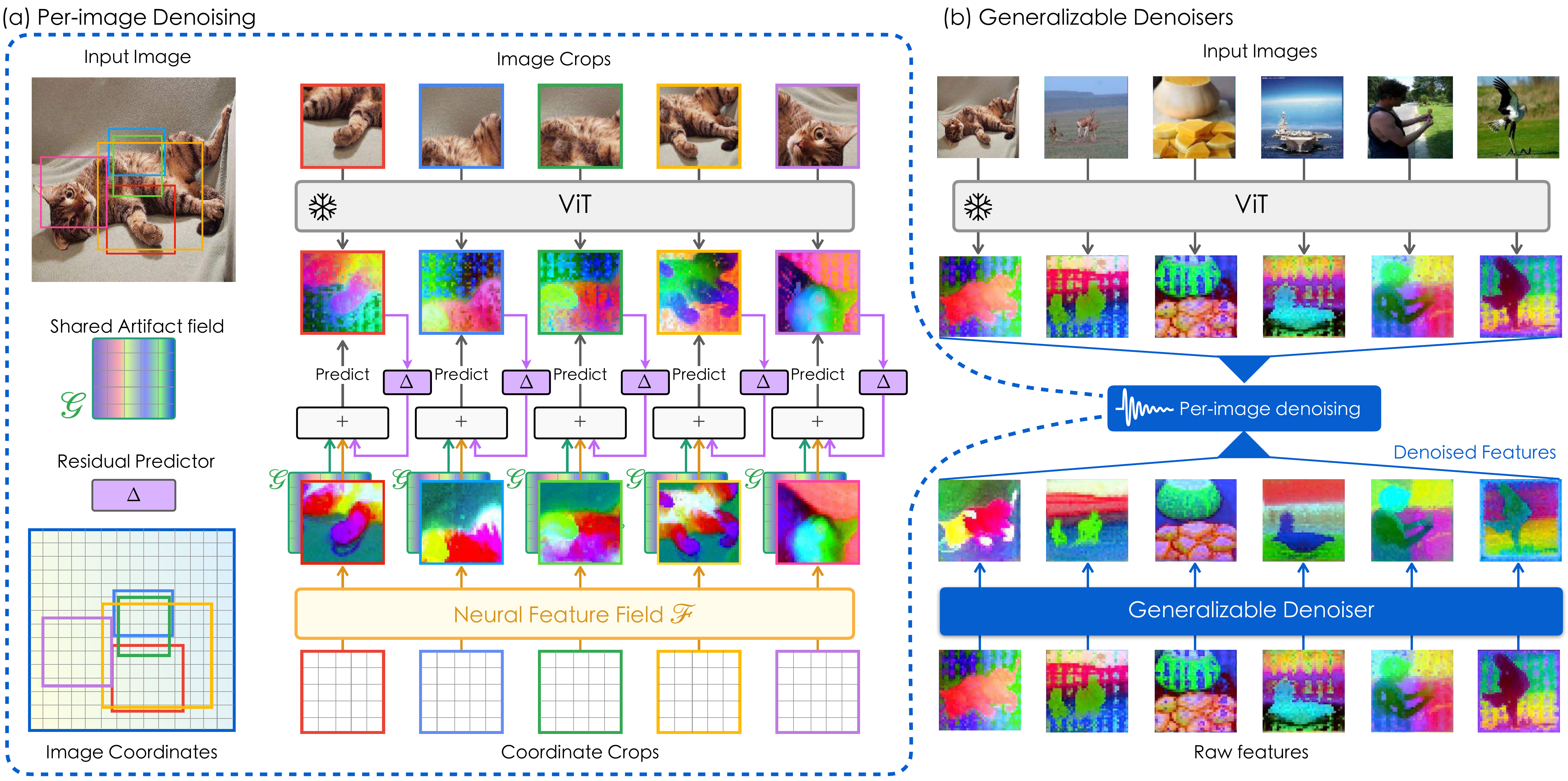}
    \caption{\textbf{Method Overview.} \ours consists of a two-stage denoising pipeline. (a) In the first stage, our method decomposes the raw feature of an image crop into a noise-free semantics term $\mathcal{F}$, an input-independent, position-related artifact term $\mathcal{G}$, and an additional residual term $\Delta$. (b) In the second stage, we train a generalizable denoiser to predict clean features from their original features. At inference time, only a single feedforward is needed to obtain denoised features.}
    \label{fig:method}
\end{figure}

In this section, we start by analyzing ViT outputs to motivate our approach (\S\ref{sec:factorization}). Then, we introduce our per-image denoising method, which removes artifacts and produces noise-free features (\S\ref{sec:per_img_denoising}). Lastly, we explain how the noise-free features are utilized as pseudo-labels to train a generalizable denoiser (\S\ref{sec:gen_denoiser}). Our method pipeline is depicted in \cref{fig:method}.

\subsection{Factorizing ViT Outputs}
\label{sec:factorization}
Our method is grounded in the principle that ideal visual features should be inherently translation and reflection invariant, \ie, the features of an object should remain consistent, regardless of changes in the viewing window, size, and orientation. However, as indicated in \cref{eq1,eq2,eq3,eq4} and \cref{fig:vit_arch}-(b), ViTs intermix patch embeddings with positional embeddings, thereby breaking the transformation invariance of visual features. This breach of invariance might not appear immediately problematic, but our investigations, illustrated in \cref{fig:vit_arch}-(a) and (c), reveal a distinct correlation between the inclusion of positional embeddings and the emergence of undesirable artifacts in ViT outputs. Particularly, the middle row of \cref{fig:vit_arch}-(c) shows that these artifacts persist with minor variation across different images, highlighting their consistency independent of the input content.

These observations motivate us to decompose ViT outputs into three terms: (1) an input-dependent, noise-free semantics term $f(\vb{x})$\footnote{Throughout this paper, we use ``noise'' and ``artifact'' interchangeably.}; (2) an input-independent artifact term related to spatial positions $g(\vb{E}_{pos})$; (3) and a residual term that accounts for the interdependence of semantics and positions  $h(\vb{x}, \vb{E}_{pos})$. The decomposition is formally expressed as:
\begin{equation}
    \mathrm{ViT}(\vb{x}) \approx f(\vb{x}) + g(\vb{E}_{pos}) + h(\vb{x}, \vb{E}_{pos}) \label{eq:decomposition}
\end{equation}

This factorization is universally applicable to all ViTs. For example, in scenarios where the output feature map is spatially invariant (\eg, no positional embedding is used), the sum of $g$ and $h$ becomes a constant bias term that can be merged into $f$.

\subsection{Per-image Denoising with Neural Fields}
\label{sec:per_img_denoising}
Directly addressing the above decomposition problem within a single forward pass in a ViT is impractical due to the intertwined nature of output features. To overcome this, we exploit the consistencies in cross-view features and artifacts: (1) \textit{Feature consistency} refers to the transformation invariance of visual features, where despite the varied spatial transformations, the semantic content remains invariant; (2) \textit{Artifact consistency} means that the input-independent artifact remains observable and constant across all transformations. Formally, consider an image $\vb{x}$ and a set of its randomly transformed views $T(\vb{x})=\{t_0(\vb{x}), t_1(\vb{x}), \cdots\}$, where each transformation $t_{i}$ is drawn from a distribution of random augmentations $\mathcal{T}$, consisting of random resizing, cropping, and flipping. Our goal is to derive a mapping $f$ such that the semantic features obtained from any transformed view, $f\left(t\left(\vb{x}\right)\right)$, are equivalent to the transformed original semantic features, $t\left(f(\vb{x})\right)$; that is, $f\left(t\left(\vb{x}\right)\right)=t\left(f(\vb{x})\right)$ with $t \sim \mathcal{T}$. Next, we describe our approach for learning the different terms in \cref{eq:decomposition} in conjunction to derive $f$.

\paragraph{Neural fields as feature mappings.} At the core of our approach is to have a holistic image semantics representation $\mathcal{F}$, for \textit{each individual image}, alongside a spatial artifact feature representation, $\mathcal{G}$, shared by \textit{all transformed views}. The holistic image feature representation $\mathcal{F}$ is designed to capture spatially independent, artifact-free semantics, while $\mathcal{G}$ should encode position-dependent but input-independent noise. We use coordinate networks, known as neural fields \cite{tancik2020fourfeat,muller2022instant,shen2023distilled,kobayashi2022decomposing,kerr2023lerf,yang2023emernerf}, to actualize $\mathcal{F}$ and $\mathcal{G}$. Specifically, we define $f(t(\vb{x}))=\mathcal{F}(\mathrm{coords}(t(\vb{x})))$, where $\mathrm{coords}(\cdot)$ extracts the pixel coordinates of the transformed views relative to the original image $\vb{x}$, and $g(\vb{E}^{i}_{pos})=\mathcal{G}(i)$, with $i \in \{0, \cdots, N-1\}$ denoting the patch index. For simplicity, we use $\mathcal{G}$ to denote the 2D artifact feature map reshaped from the 1D ordered sequence $\{\mathcal{G}(i)\}_{i=0}^{N-1}$. We refer to $\mathcal{F}$ and $\mathcal{G}$ as the semantics field and the artifact field, respectively.

\paragraph{Learning the decomposition.} We learn the semantics field $\mathcal{F}$, the artifact field $\mathcal{G}$, and the residual term $\Delta$ by minimizing a regularized reconstruction loss:
\begin{align}
    \mathcal{L}_\text{recon}      & = \mathcal{L}_{\text{distance}} + \alpha \mathcal{L}_{\text{residual}} + \beta  \mathcal{L}_{\text{sparsity}} \label{eq:recon}                             \\
    \mathcal{L}_{\text{distance}} & = 1 - \cos(\vb{y}, \widehat{\vb{y}}) + \| \vb{y} - \widehat{\vb{y}} \|_2,                                                                                  \\
    \mathcal{L}_\text{residual}   & = \|\mathrm{sg}\left(\vb{y} - \widehat{\vb{y}'}\right) - \widehat{\Delta}\|_2, \hspace{1em} \mathcal{L}_{\text{sparsity}} = \|\widehat{\Delta}\|_1         \\
    \text{where}~~\vb{y}          & = \mathrm{sg}\left(\mathrm{ViT}\left(t\left(\vb{x}\right)\right)\right), \hspace{2em} \widehat{\vb{y}} = \widehat{\vb{y}'} + \mathrm{sg}(\widehat{\Delta}) \\
    \widehat{\vb{y}'}             & = \mathcal{F}_{\theta}(\mathrm{coords}(t(\vb{x}))) + \mathcal{G}_{\xi}, \hspace{0.6em} \widehat{\Delta} = h_{\psi}(\vb{y})
\end{align}
Here, $\cos(\cdot, \cdot)$ denotes the cosine similarity, $\mathrm{sg}(\cdot)$ represents the stop-gradient operation, $t(\cdot)$ is a random transformation sampled from $\mathcal{T}$, and $\theta$, $\xi$ and $\psi$ are the learnable parameters. Our loss function is designed to encourage $\widehat{\Delta}$ to remain minimal by imposing a sparsity regularization, thereby allowing $\widehat{\vb{y}'}$ to represent as much of the ViT output as possible. The use of stop-gradient operators is to avoid trivial solutions, such as identity mapping. The reconstructed feature from our method is $\widehat{\vb{y}} = \mathcal{F}_{\theta}\left(\mathrm{coords}\left(t\left(\vb{x}\right)\right)\right) + \mathcal{G}_{\xi} + \mathrm{sg}\left(h_{\psi}\left(\mathrm{ViT}\left(t\left(\vb{x}\right)\right)\right)\right)$, each term corresponding to $f, g$, and $h$ as defined in \cref{eq:decomposition}.

\paragraph{Optimization.} We break our optimization process into two phases, each spanning half of the total training iterations. In the first phase, we train $\mathcal{F}_\theta$ and $\mathcal{G}_{\xi}$ using only $\mathcal{L}_{\text{distance}}$, allowing them to capture a significant portion of the ViT outputs. After completing half of the optimization iterations, we freeze $\mathcal{G}_\xi$ and continue to train $\mathcal{F}_\theta$ alongside $h_\psi$ using $\mathcal{L}_\text{recon}$ for the rest iterations. The coefficients $\alpha$ and $\beta$ in $\mathcal{L}_\text{recon}$ balance loss scales and regulate the residual term to prevent $\widehat{\Delta}$ from over-explaining the outputs.

\subsection{Generalizable Denoiser}
\label{sec:gen_denoiser}
Our per-image denoising method can already effectively remove artifacts from ViT outputs, yielding visually stunning denoised feature maps. The problems we are left with are run-time efficiency and distribution shifts. Specifically, the per-image denoising process is suitable for offline applications but undesired for real-time applications, and individually denoised feature maps can lead to feature distribution shifts due to sample bias, which hampers the feature coherence across images. To address these issues, we introduce a generalizable denoiser.

After applying per-image denoising, we accumulate a dataset of pairs consisting of noisy ViT outputs $\vb{y}$ and their denoised counterparts $\mathcal{F}$, denoted as $\mathcal{B}=\{\left(\vb{y}_i, \mathcal{F}_i\right)\}_{i=1}^B$. We then train a denoiser network  ${D}_\zeta$ to predict noise-free features from raw ViT outputs, \ie, $\hat{\mathcal{F}} = D_\zeta(\vb{y})$. The loss function is:
\begin{align}
    \mathcal{L}_{\text{distance}}^\text{\ours} & = 1 - \cos\left(D_\zeta\left(\vb{y}\right), \mathcal{F}\right) + \|D_\zeta\left(\vb{y}\right) - \mathcal{F} \|_2 \label{eq:gen_loss}
\end{align}
Our generalizable denoiser is implemented as a single Transformer block, supplemented with additional learnable positional embeddings that are applied post the forward pass of a ViT. This design aims to mitigate the input-independent artifacts. To predict denoised features, the outputs from a pre-trained ViT are added with these positional embeddings and then processed through the Transformer block.

Notably, this learned denoiser is lightweight, thus adding negligible latency to the original ViT and facilitating real-time applications. It also learns to generalize across samples, mitigating the distribution shift issue in the per-image denoising process.

\section{Experiments}
\label{sec:experiments}

In this section, we first explore if ViTs trained with different objectives all have artifacts. Then, we evaluate the effectiveness of our generalizable denoiser on dense prediction tasks. For all experiments, we default to using ViT-\textit{base} models with patch sizes of 14 or 16, depending on the availability of their implementations and model weights in PyTorch Image Models (\texttt{timm} \cite{rw2019timm}). We defer all the implementation details to the appendix.

\subsection{Artifacts in ViTs}

\paragraph{Positional artifacts in different ViTs.}
We visualize feature maps from differently pre-trained ViTs in \cref{fig:teaser}. Among these, DINOv2 \cite{oquab2023dinov2}, a state-of-the-art vision foundation model with excellent performance on downstream tasks, displays clear position-related artifacts. Additionally, DeIT-III \cite{touvron2022deit}, trained with image class labels, and CLIP \cite{radford2021learning}, trained by text-image alignment, also exhibit noticeable artifacts. Furthermore, EVA02 \cite{fang2023eva}, which distills local patch features from a pre-trained CLIP model using masked image modeling, also has clear feature artifacts. In ViTs we have tested, our proposed \ours successfully mitigates these artifacts (``Original features'' \textit{vs.} ``Denoised features'' in \cref{fig:teaser}).

\begin{figure}[t!]
    \centering
    \includegraphics[width=1\linewidth]{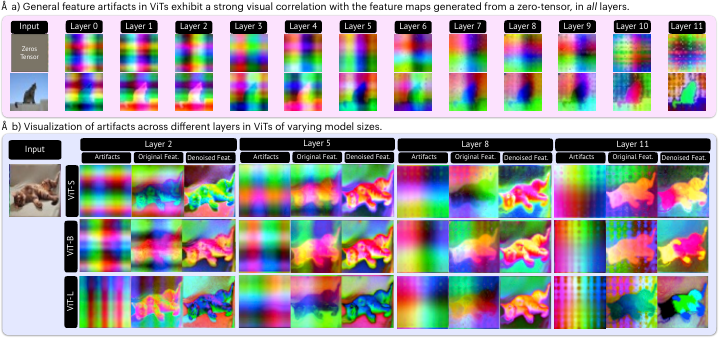}
    \caption{\textbf{Visual analysis of ViT output features and denoised features.} (a) Visualizations of the feature maps from all layers of a DINOv2 \cite{oquab2023dinov2} ViT-\textit{base} model. Notably, the artifacts in the feature maps derived from the cat image exhibit a strong visual correlation with those from the zero-tensor inputs. (b) Visualizations of the decomposed artifacts, the original features, and the denoised features across various layers of DINOv2 ViTs. We observe similar patterns in differently-sized models.}
    \label{fig:multilayer}
\end{figure}

\paragraph{Artifacts in different layers.} In \cref{fig:multilayer}, we present a visual analysis of the artifact decomposition across various layers of DINOv2 ViTs of different sizes (b), alongside feature maps generated using only zero-tensors as input (a). Notably, the artifacts decomposed by our \ours show a strong visual resemblance to these zero-tensor-input feature maps. In addition, we observe that the artifacts vary across layers: the shallower layers predominantly exhibit low-frequency patterns, whereas the deeper layers are characterized by high-frequency patterns. Importantly, these patterns are consistent across ViTs of different sizes (\eg, from ViT-\textit{small} to ViT-\textit{large}), diverging from the hypothesis in \cite{darcet2023vision} that only large ViTs would display such patterns.

\begin{table*}[t!]
    \centering
    \small
    \caption{\textbf{Comparison of features correlation to spatial positions.} We report the maximal information coefficient (MIC) between grid features and their coordinates.}
    \tablestyle{3.5pt}{1.2}
    \begin{tabular}{l | x{60} | x{60} | x{60}}
                                        & \textbf{Before denoising} & \multicolumn{2}{c}{\textbf{After denoising}}                                         \\

        Method                          & \textbf{Raw features}     & \textbf{Artifacts term $\mathcal{G}$}        & \textbf{Semantics term $\mathcal{F}$} \\
        \shline
        DINOv2 \cite{oquab2023dinov2}   & 0.44                      & 0.54                                         & 0.22                                  \\
        DeiT-III \cite{touvron2022deit} & 0.34                      & 0.32                                         & 0.06                                  \\
        CLIP \cite{radford2021learning} & 0.11                      & 0.14                                         & 0.08                                  \\
    \end{tabular}
    \label{tab:correlation}
\end{table*}

\paragraph{Correlation between artifacts and positions.} Beyond visual qualitative inspection, we aim to quantitatively analyze the correlation between artifacts and their positions. Similar to \cite{voita2023neurons}, we use the maximal information coefficient (MIC) to measure the dependency between grid features and their normalized patch coordinates (See appendix for more details). This metric indicates how much patch features depend on their spatial positions and semantic content. As shown in \cref{tab:correlation}, both the original ViT outputs and the decomposed artifacts exhibit a higher spatial correlation than the denoised semantic features, irrespective of the training methodology employed. These results support our hypothesis about the significant role of positional embeddings in the emergence of artifacts. Note that there is no ``ground-truth'' quantitative metric to to definitively quantify these patterns; hence, our reported numerical results should be viewed as empirical indicators, akin to the ``high-norm'' indicator used in \cite{darcet2023vision}.

\subsection{Evaluation on Downstream Task Performance}

We evaluate our method in dense recognition tasks, including semantic segmentation, monocular depth estimation, object detection, and object discovery. It is important to note that there is no direct competitor for these tasks in our study. Instead, our focus is on comparing the performance of pre-trained ViTs before and after applying our \ours. For all the models in the main experiments, we use 10k denoised samples randomly selected from the VOC2012 and the VOC2007 datasets, excluding their validation samples, to train generalizable denoisers.

\paragraph{Semantic segmentation.} We follow \cite{oquab2023dinov2,darcet2023vision} to evaluate our approach in two semantic segmentation datasets: VOC2012 \cite{pascal-voc-2012}
and ADE20k \cite{zhou2018semantic}, using a linear probing protocol, \ie, a linear layer is trained to predict pixels' class from patch tokens.
\cref{tab:dense_results}
presents the main results. We observe significant and consistent enhancements in all pre-trained ViTs across datasets. Notably, the DINOv2-\textit{giant}, with an 83.0 mIoU on VOC2012 as reported in \cite{oquab2023dinov2}, is outperformed by our \ours-denoised DINOv2-\textit{base} model (84.84 mIoU). This improvement is also evident in the ADE20k dataset, where the DINOv2-\textit{giant} and DINOv2-\textit{large} models attain mIoUs of 49.0 and 47.7, respectively, as reported in \cite{oquab2023dinov2}, while our denoised \textit{base} model achieves a 48.66 mIoU. Remarkably, the \textit{giant} model, which is $\mathbf{13\times}$ larger than the \textit{base} model, is outperformed by or on par with our denoised \textit{base} model. This indicates that the performance gains primarily stem from effective artifact removal rather than the \textit{minor} increase in model parameters of our denoiser network.

Our \ours also increases the performance of the concurrent DINOv2-reg model \cite{darcet2023vision}, where a ViT is trained with dummy learnable register tokens. As evidenced in \cref{tab:dense_results}, our \ours enhances the performance of both DINOv2 ((f1) \textit{vs.} (f2)) and DINOv2-reg ((e1) \textit{vs.} (e2)). When applying \ours only, DINOv2 shows more improvements compared to using registers ((f2) \textit{vs.} (e1)); for instance, DINOv2 denoised by \ours achieves 84.84 mIoU in VOC2012 and 48.66 mIoU in ADE20k, surpassing the performance of DINOv2-reg, which achieves 83.64 mIoU and 48.22 mIoU on the respective benchmarks. Furthermore, \ours can further enhance the performance of DINOv2-reg ((e1) \textit{vs.} (e2)) on both datasets (+0.86 in VOC2012 and +1.12 in ADE20k).
In addition, DINOv2-reg \cite{darcet2023vision} requires retraining entire models from scratch using 142M images, while our approach requires training a single Transformer block using 10k denoised samples.

\begin{table*}[t!]
    \centering
    \small
    \caption{\textbf{Quantitative performance of \ours.} \ours improves differently pre-trained ViTs for dense prediction tasks. We report performance on semantic segmentation (VOC2012, ADE20K) and depth prediction (NYUd) tasks.}
    \tablestyle{3.5pt}{1.2}
    \begin{tabular}{l | x{35}x{35} | x{35}x{35} | x{35}x{35} }
                                                                 & \multicolumn{2}{c|}{VOC2012 Segmentation} & \multicolumn{2}{c|}{ADE20k Segmentation} & \multicolumn{2}{c}{NYUv2 Depth Estimation}                                                     \\
        Method
                                                                 & mIoU($\uparrow$)
                                                                 & mAcc($\uparrow$)
                                                                 & mIoU($\uparrow$)
                                                                 & mAcc($\uparrow$)                          & $\delta_1$($\uparrow$)                   & abs rel($\downarrow$)                                                                          \\
        \shline
        (a1) DINO \cite{caron2021emerging}                       & 63.00                                     & 76.35                                    & 31.03                                      & 40.33          & 73.19          & \textbf{0.1701} \\
        (a2) DINO \cite{caron2021emerging} + \textbf{\ours}      & \textbf{66.22}                            & \textbf{78.14}                           & \textbf{32.40}                             & \textbf{42.01} & \textbf{73.53} & 0.1731          \\
        \hline
        (b1) DeiT-III \cite{touvron2022deit}                     & 70.62                                     & 81.23                                    & 32.73                                      & 42.81          & \textbf{72.16} & \textbf{0.1788} \\
        (b2) DeiT-III \cite{touvron2022deit} + \textbf{\ours}    & \textbf{73.36}                            & \textbf{83.74}                           & \textbf{36.57}                             & \textbf{49.01} & 71.36          & 0.1802          \\
        \hline
        (c1) EVA02 \cite{fang2023eva}                            & 71.52                                     & 82.95                                    & 37.45                                      & 49.74          & 63.68          & 0.1989          \\
        (c2) EVA02 \cite{fang2023eva}  + \textbf{\ours}          & \textbf{73.15}                            & \textbf{83.55}                           & \textbf{37.87}                             & \textbf{49.81} & \textbf{68.52} & \textbf{0.1964} \\
        \hline
        (d1) CLIP \cite{radford2021learning}                     & 77.78                                     & 86.57                                    & 40.51                                      & 52.47          & 73.95          & 0.1679          \\
        (d2) CLIP \cite{radford2021learning} + \textbf{\ours}    & \textbf{79.01}                            & \textbf{87.48}                           & \textbf{41.10}                             & \textbf{53.07} & \textbf{74.61} & \textbf{0.1667} \\
        \hline
        (e1) DINOv2-reg \cite{darcet2023vision}                  & 83.64                                     & 90.67                                    & 48.22                                      & 60.52          & 87.88          & 0.1190          \\
        (e2) DINOv2-reg \cite{darcet2023vision} + \textbf{\ours} & \textbf{84.50}                            & \textbf{91.45}                           & \textbf{49.34}                             & \textbf{61.70} & \textbf{88.26} & \textbf{0.1157} \\
        \hline
        (f1) DINOv2 \cite{oquab2023dinov2}                       & 83.60                                     & 90.82                                    & 47.29                                      & 59.18          & 86.88          & 0.1238          \\
        (f2) DINOv2 \cite{oquab2023dinov2} + \textbf{\ours}      & \textbf{84.84}                            & \textbf{91.70}                           & \textbf{48.66}                             & \textbf{60.24} & \textbf{87.58} & \textbf{0.1200}
    \end{tabular}
    \label{tab:dense_results}
\end{table*}

\paragraph{Depth estimation.} Following \cite{oquab2023dinov2}, we evaluate our method on the NYUv2-Depth dataset \cite{Silberman_ECCV12} using a linear evaluation protocol (more details in appendix). As shown in \cref{tab:dense_results}, our method clearly enhances the performance of most pre-trained ViTs. For context, the DINOv2-\textit{large} model exhibits a 0.01 RMSE improvement over the DINOv2-\textit{base} model with 3.5$\times$ more parameters. Our denoiser achieves similar performance gains with 0.08$\times$ the parameters of the base model. These results highlight our method's efficiency, achieving marked performance gains with minimal increases in parameter count.

\begin{table}[t!]
    \centering
    \small
    \caption{\textbf{Object detection with frozen features.} We report the mAP metric on the VOC object detection benchmark.}
    \tablestyle{3.5pt}{1.2}
    \begin{tabular}{l | y{40} y{50} y{40}  y{40}  y{40} y{40}}
        Method                               & DINOv2 \cite{oquab2023dinov2}
                                             & DINOv2-reg \cite{darcet2023vision}
                                             & CLIP \cite{radford2021learning}
                                             & DeiT-III \cite{touvron2022deit}
                                             & DINO \cite{caron2021emerging}
                                             & EVA02 \cite{fang2022eva}                                                                                                                                                                                                                                                    \\
        \shline
        baseline                             & 81.4                                       & 80.9                                       & 80.9                                       & 75.8                                       & 76.4                                       & 79.4                                       \\
        \baseline{baseline + \textbf{\ours}} & \baseline{\textbf{81.9} \increase{(+0.5)}} & \baseline{\textbf{81.4} \increase{(+0.5)}} & \baseline{\textbf{81.7} \increase{(+0.9)}} & \baseline{\textbf{77.0} \increase{(+1.2)}} & \baseline{\textbf{77.1} \increase{(+0.7)}} & \baseline{\textbf{80.2} \increase{(+0.8)}} \\
    \end{tabular}
    \label{tab:obj_det}
\end{table}

\paragraph{Object detection.} In this experiment, we train ViTDet detectors \cite{li2022exploring} on the frozen features following the Faster RCNN framework \cite{ren2015faster} (more details in appendix). We train all models on the VOC \texttt{trainval07+12} subset and report their mAP metrics on the \texttt{test2007} subset. Results are reported in \cref{tab:obj_det}. Our approach shows consistent improvements over the studied ViTs. Notably, DINOv2-reg \cite{darcet2023vision} shows a slight decrease in object detection performance when compared to the original DINOv2 \cite{oquab2023dinov2}, while our approach improves it.

\begin{figure}[t!]
    \centering
    \includegraphics[width=1\linewidth]{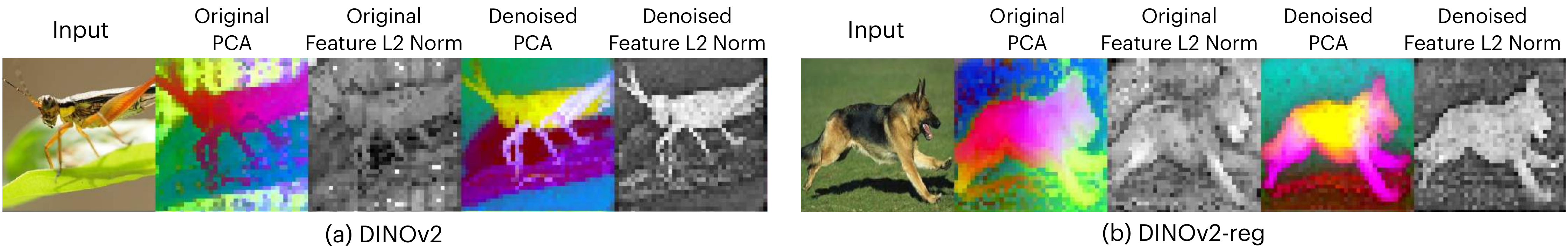}
    \caption{\textbf{Emerged object discovery ability.} The features denoised by our \ours show higher feature norms on objects of interest.}
    \label{fig:emerg_obj_dis}
\end{figure}

\begin{table*}[t!]
    \centering
    \caption{\textbf{Unsupervised object discovery using LOST \cite{simeoni2021localizing}.} We report the corloc score across three datasets. Our \ours significantly improves existing models. $^\dagger$: results quoted from \cite{darcet2023vision}; these models are ViT-\textit{large} trained on the ImageNet-22k dataset while our reported results are based on the publicly available ViT-\textit{base}.}
    \tablestyle{3.5pt}{1.2}
    \begin{tabular}{l | y{50} y{50} y{50}}
        Method                                                & VOC2007                                     & VOC2012                                     & COCO20k                                     \\
        \shline
        (a) \gc{$^\dagger$DINOv2} \cite{oquab2023dinov2}   & \gc{35.3}                                   & \gc{40.2}                                   & \gc{26.9}                                   \\
        (b) \gc{$^\dagger$DINOv2-reg} \cite{darcet2023vision} & \gc{55.4}                                   & \gc{60.0}                                   & \gc{42.0}                                   \\
        \hline
        (c) DINOv2-reg                                        & 38.0                                        & 41.5                                        & 26.9                                        \\
        \baseline{(d) DINOv2-reg + \textbf{\ours}}            & \baseline{\textbf{56.1} \increase{(+18.1)}} & \baseline{\textbf{59.3} \increase{(+17.8)}} & \baseline{\textbf{45.5} \increase{(+18.6)}} \\
        \hline
        (e) DINOv2                                            & 30.8                                        & 35.9                                        & 23.4                                        \\
        \baseline{(f) DINOv2 + \textbf{\ours}}                & \baseline{\textbf{58.0} \increase{(+27.2)}} & \baseline{\textbf{60.3} \increase{(+24.4)}} & \baseline{\textbf{46.7} \increase{(+23.3)}} \\
    \end{tabular}
    \label{tab:obj_discovery}
\end{table*}

\paragraph{Object discovery.} Unsupervised object discovery has been a long-standing problem of interest. An intriguing finding from our experiments is the emerging capability of object discovery in denoised ViTs. \cref{fig:emerg_obj_dis} illustrates this through PCA visualizations and $L2$ norms of the feature maps. Post-denoising, not only are the artifacts removed, but also the objects of interest become more distinctly visible from the feature norm values. This enhancement in object clarity is \textit{not} a goal of \ours but emerges as the outcome of our method.

To quantitatively assess these enhancements, we follow \cite{darcet2023vision} to use LOST \cite{simeoni2021localizing} for evaluating object discovery efficacy before and after applying our \ours. We use feature norms as an indicator of object prominence. We conduct object discovery experiments on PASCAL VOC 2007 \cite{pascal-voc-2007} and 2012 \cite{pascal-voc-2012} and COCO20k datasets \cite{lin2014microsoft}.
\cref{tab:obj_discovery} presents the results. Our \ours significantly improves both DINOv2 \cite{oquab2023dinov2} and DINOv2-reg \cite{darcet2023vision} in all the evaluated datasets. In particular, while the publicly available DINOv2-reg shows some improvements ((c) \textit{vs.} (e)), we find that it falls short of the performance levels reported in \cite{darcet2023vision} ((c) \textit{vs.} (b)). Despite this, our \ours achieves more substantial enhancements in object discovery capabilities, even compared to the numbers reported in \cite{darcet2023vision} ((f) \textit{vs.} (b)).

\noindent
{
    \begin{minipage}[t]
        {0.41\textwidth}
        \centering
        \small
        \makeatletter\def\@captype{table}\makeatother\caption{Ablation study on per-image denoising using KNN segmentation protocol on VOC12\texttt{val}.}
        \tablestyle{3.5pt}{1.2}
        \begin{tabular}{l | x{40}}
            Representations                                           & mIoU                      \\
            \shline
            (a) DINOv2                                                & 65.35                     \\
            \hline
            (b) $\mathcal{F}$                                         & 67.81                     \\
            (c) $\mathcal{F} + \mathcal{G}$                           & 70.82                     \\
            \baseline{(d) $\mathcal{F} + \mathcal{G} + \hat{\Delta}$} & \baseline{\textbf{70.94}} \\
        \end{tabular}
        \label{tab:per_img_denoising}
    \end{minipage}
    \quad
    \begin{minipage}[t]{0.53\textwidth}
        \centering
        \small
        \makeatletter\def\@captype{table}\makeatother\caption{Ablation study on the architectural design of generalizable denoiser. We report the mIoU of the VOC2012 validation set.}
        \tablestyle{3.5pt}{1.2}
        \begin{tabular}{l | x{40}}
            Denoiser architectures                        & mIoU                      \\
            \shline
            (a) DINOv2 (reproduced)                       & 83.60                     \\
            \hline
            (b) conv1x1                                   & 82.15                     \\
            (c) conv3x3                                   & 83.27                     \\
            \baseline{(d) Single Transformer Block + PE.} & \baseline{\textbf{84.84}} \\
            (e) Single Transformer Block                  & 84.81                     \\
        \end{tabular}
        \label{tab:denoiser}
    \end{minipage}
}

\subsection{Ablation Study}
In this section, we provide ablation studies to understand the importance of different components in our proposed \ours.

\paragraph{Factorization.} We ablate our per-image denoising method using a K-Nearest-Neighbor (KNN) pixel segmentation evaluation protocol on the VOC2012 dataset. Specifically, we collect class centroids from each training image by masked pooling to construct a memory bank using ground truth annotations. Then, for each pixel in a validation image, we classify it based on its 20 nearest neighbors in the memory bank. We report the mIoU on the validation set. \cref{tab:per_img_denoising} shows the results. We observe that combining the artifact field $\mathcal{G}$ and the residual term $\hat{\Delta}$ yields the best result (d). Omitting both these elements reduces our approach to merely utilizing a neural field $\mathcal{F}$ to learn multi-crop ensembled image features, without addressing artifacts (b). While this variant shows improvement, it falls behind our proposed method by a large margin, underscoring the importance of removing artifacts.

\paragraph{Generalizable denoiser.} We explore alternative architectural designs for our generalizable denoiser in \cref{tab:denoiser}. We study four variations: 1) our default setting, which incorporates a single Transformer Block with new learnable position embeddings; 2) our default setting but without position embeddings; 3) a multi-layer convolution denoiser with a \texttt{Conv1x1-ReLu-Conv1x1-ReLu-Conv1x1} structure, and 4) a multilayer convolution denoiser with a \texttt{Conv3x3-ReLu-Conv3x3-ReLu-Conv3x3} structure. We observe that denoisers based on convolutional structures (b, c) do not yield good results, with the conv1x1 setting performing the worst (c).
Moreover, we note that our default setting with a Transformer block and learnable positional embeddings achieves the best result (d), and removing the learnable position embeddings obtains very similar numerical performance (e). We empirically find that the design of (d) leads to better qualitative visualizations, and thus we use this setting.

\begin{figure}[t!]
    \centering
    \includegraphics[width=1\linewidth]{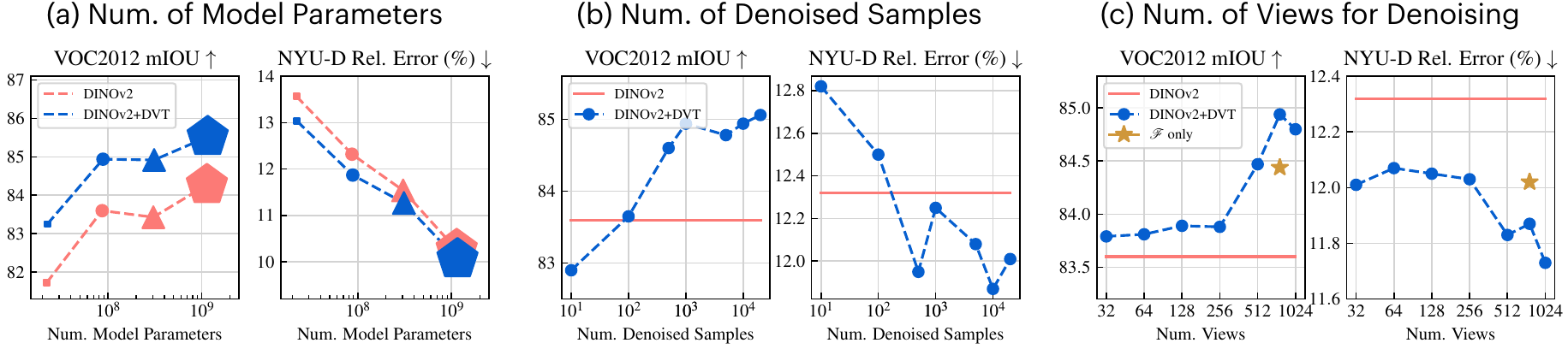}
    \caption{\textbf{DVT's Scaling Behaviors.} We study the generalizable denoiser's performance for (a) different model sizes, (b) the number of denoised samples used for training denoisers, and (c) the number of views used when performing per-image denoising.}
    \label{fig:scaling}
\end{figure}

\paragraph{Scaling behaviors.} We study how \ours scales with model sizes and data scales in \cref{fig:scaling}. In (a), we see \ours boosts differently-sized ViTs, even allowing ViT-\textit{base} to match or exceed ViT-\textit{giant} performance in semantic segmentation. Overall, \ours's scaling behaviors closely align with those of baseline models. In (b), we study the impact of the number of denoised training samples on task performance, where \ours shows promising results even with limited training samples (\eg, 100$\sim$1000). Note that our denoiser never sees ADE20k and NYU-depth datasets during training, yet generalizes effectively.
In (c), we plot the task performance \vs the number of views used for the denoising. \ours benefits from more views in first-stage denoising. When training neural fields, more views enhance performance, while fewer views lead to overfitting.
In particular, aggregating views is itself an approach to denoising, which still \textit{aligns with} our motivation.
We also demonstrate that a denoiser trained on samples denoised solely by aggregating views via neural fields (${\mathcal{F}}$-only in (c)) surpasses baselines but underperforms the full \ours, which further confirms the effectiveness of our proposed denoising procedure.

\section{Discussion and Future Works}
\label{sec:discussion}

\ourslong (\ours) introduces a robust method leveraging neural fields to eliminate feature artifacts from ViTs. This work additionally pinpoint positional embeddings as the primary source of these artifacts, despite their importance in various vision tasks. Using a neural field optimization process, \ours efficiently extracts clean features from the noise-riddled feature maps of existing ViTs. And using a scalable feature denoiser model, \ours eliminates the need for individual image optimizations. When learned from individually denoised samples, our denoiser generalizes well to unseen data and improves pre-trained ViTs by large margins in dense vision tasks. More broadly, our research suggests several avenues for future exploration: (1) understanding the role of positional embeddings in ViT could inform the design of next-generation deep learning architectures, and (2) redefining positional embeddings within ViTs and transformers is also an imperative problem. Lastly, combining the insights from our work and those of \cite{darcet2023vision} could lead to a \textit{more complete} picture of how these artifacts emerge. We hope that the results presented in this work contribute to a deeper understanding of artifacts in vision transformers and beyond.

\section*{Acknowledgements}
We are grateful to many friends,
including Jiageng Mao, Junjie Ye, Justin Lovelace, Varsha Kishore, and Christian Belardi, for their fruitful discussions on this work and follow-ups. Katie Luo is supported by an Nvidia Graduate Fellowship. Leonidas Guibas acknowledges the support from a Vannevar Bush Faculty Fellowship. We also acknowledge an unrestricted gift from Google in support of this project.


\bibliographystyle{splncs04}
\bibliography{main}

\clearpage
\vbox{

    \appendix

    \centering
    {\Large\bf
        Supplementary Material: \\
        \ourslong
        \par}
}
\vspace{10pt}

\renewcommand{\thesection}{\Alph{section}}
\renewcommand{\thesubsection}{\thesection.\arabic{subsection}}
\renewcommand{\thetable}{S\arabic{table}}
\renewcommand{\thefigure}{S\arabic{figure}}
\renewcommand{\theequation}{S\arabic{equation}}
\setcounter{figure}{0}
\setcounter{table}{0}
\setcounter{equation}{0}

In the appendix, we provide detailed implementation details in \S\ref{sec:impl_details}, elaborate on evaluation protocols and additional results in \S\ref{sec:evaluation_protos}, and discuss the understanding of position embeddings in ViT in \S\ref{sec:understanding}. Lastly, we discuss the limitations of this work and propose a few avenues for future work in \S\ref{sec:limitation}.

\section{Implementation Details}
\label{sec:impl_details}

\subsection{Denosing with Neural Fields}

Recall that we decompose the output feature map from a pre-trained ViT into three components: $\mathbf{y}\approx \mathcal{F}(\mathcal{A}) + \mathcal{G} + \mathbf{h}(\mathbf{y})$, where $\mathcal{F}$ is a feature semantic field, $\mathcal{G}$ is an artifact field, and $\vb{h}$ is a residual predictor. We describe their implementation details below.

\paragraph{Neural field $\mathcal{F}$.} To facilitate efficient learning, we use InstantNGP \cite{muller2022instant}, a type of compact and fast coordinate network, parameterized by learnable multi-level hash grids $\mathcal{H}$ and a lightweight MLP $\phi(\cdot)$, to learn $\mathcal{F}$. It takes as input a normalized 2D coordinate $(i, j)$, within the range of [0, 1], and outputs its corresponding feature vector, \ie, $\mathcal{F}(i, j) = \phi\left(\mathcal{H}(i, j)\right)$. We refer readers to \cite{muller2022instant} for a more detailed understanding of the learnable hash grids.  In our implementation, we use a hash encoding resolution that spans from $2^4$ to $2^{10}$ with 16 levels. Each hash entry has a channel size of 8. The maximum number of hash entries of each resolution is $2^{20}$. For the lightweight MLP, we use a two-layer \texttt{Linear-ReLu-Linear} structure. The hidden dimension of this MLP is half the size of the output feature dimension, which corresponds to the feature dimension of the ViT being studied (\eg, 768 for a ViT-\textit{base} and 1024 for a ViT-\textit{large}).

\begin{figure}
    \centering
    \includegraphics[width=\linewidth]{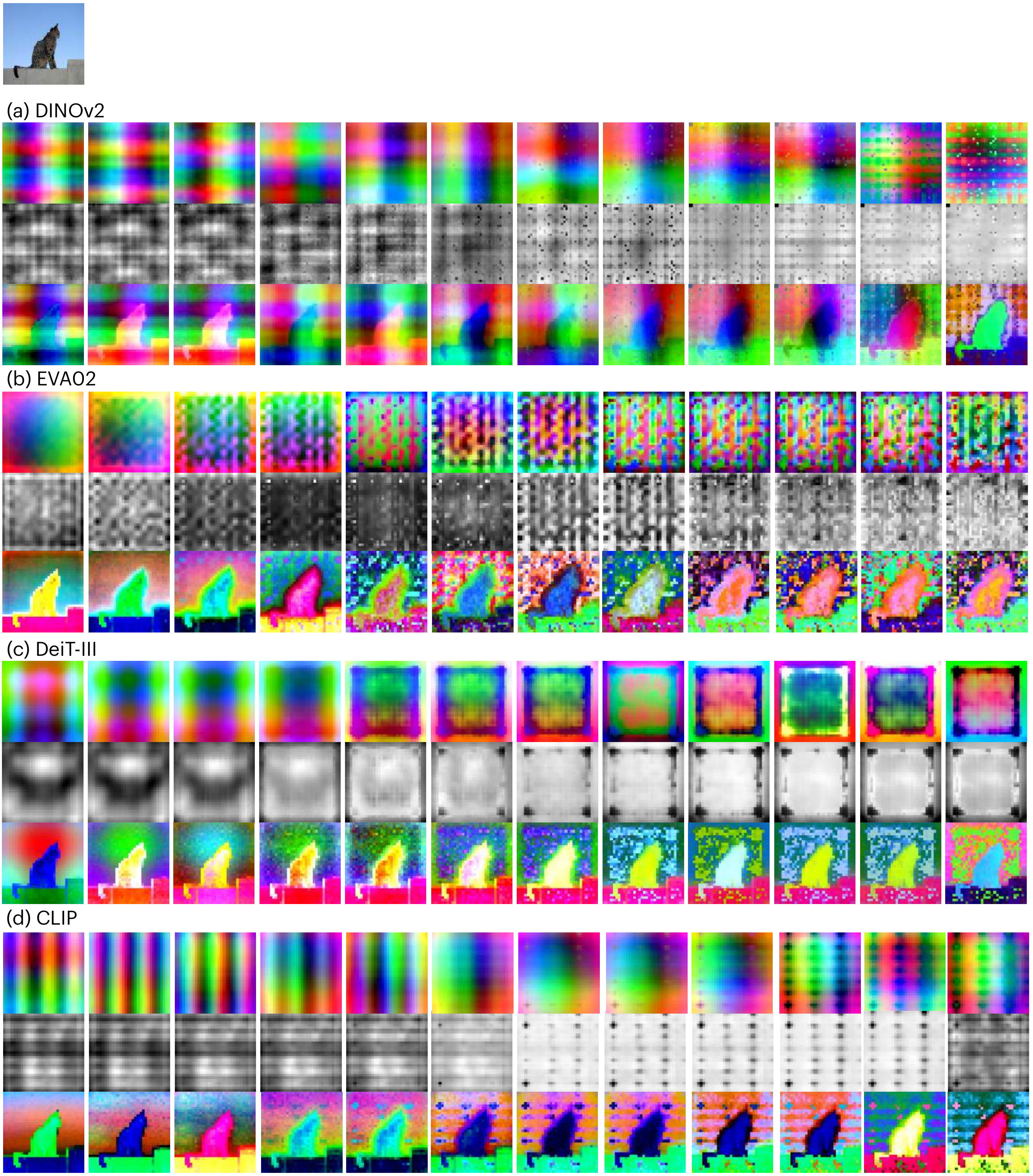}
    \caption{
        \textbf{Feature map visualizations: positional embeddings (PE) and a cat image in different ViTs.} We visualize the feature maps across different layers (1 to 12) of various pre-trained ViT-base models, displayed sequentially from left to right. For each panel, the top row shows the feature maps generated by inputting a zero tensor, highlighting the influence of PE alone. The middle row presents the feature norm of the PE feature map. The bottom row presents the feature map for a sample cat image, allowing for a comparison that reveals visual correlations between the artifacts in general image feature maps and the PE feature map.
    }
    \label{fig:strong_multilayer}
\end{figure}

\paragraph{Artifact field $\mathcal{G}$.} For all experiments, we use a 2D learnable feature map of size $C\times K \times K$ to learn the input-independent noise, where $C$ corresponds to the feature dimension of the studied ViT, and $K$ is the spatial size. We compute $K$ by $(H - P) / S + 1$, where $H$ is the height\&width of input images (which we resize to be square), $P$ is the patch size, and $S$ is the stride size used in the model. To accommodate ViTs with different patch sizes, we set $H$ to 518 for those trained with a patch size of 14, and 512 for ViTs with a patch size of 16, resulting in $K$ values of 37 and 32, respectively. Note that this feature map, $\mathcal{G}$, can be bilinearly interpolated to fit any arbitrary image resolution. We specifically choose these $K$ values to minimize the need for run-time interpolation during training, thus improving denoising efficiency.

\paragraph{Residual predictor $\mathbf{h}$.} The residual predictor is structured as a 3-layer MLP with \texttt{ReLU} activation after the hidden layers. The hidden dimension is set to be one-quarter of the channel dimension of the ViT being studied.

\paragraph{Optimization.} In our implementation, we extract  $N=768$ views (crops) from each image, applying random augmentations, which include random flipping with a probability of 0.5, and random resizing and cropping, where the size of the crop is scaled between 0.1 to 0.5 of the original image size and the aspect ratio is maintained between 3/4 and 4/3. For understanding how the number of views used for training affects the DVT's performance, please refer to \cref{fig:scaling} in the main text (our default setting is $N=768$).

The coefficients in our loss function ((\cref{eq:recon}) of the main text) are set as $\alpha=0.1$ and $\beta=0.02$. We use Adam optimizer, with a learning rate of 0.01 and a \texttt{LinearLR} decay strategy. Our models are trained for 20,000 iterations. Each iteration will process 2048 randomly sampled pixels from the pre-extracted feature maps. Note that due to the efficient implementation of $\mathcal{F}$ and the pre-extraction of patch features, our denoising typically takes about 100-160 seconds to finish (including the feature extraction time). This rapid optimization process allows us to easily amortize the denoising cost with parallel computes, thereby ensuring the practicality and applicability of our method in various scenarios.

We use the same hyperparameters for all experiments without any specific tuning. See \cref{fig:per_img_dinov2,fig:per_img_clip,fig:per_img_deit,fig:per_img_eva,fig:per_img_mae,fig:per_img_dino,fig:per_img_register} for visualizations of some examples of our per-image denoising output.

\subsection{Generalizable Denoiser}

\begin{figure}
    \centering
    \includegraphics[width=1\linewidth]{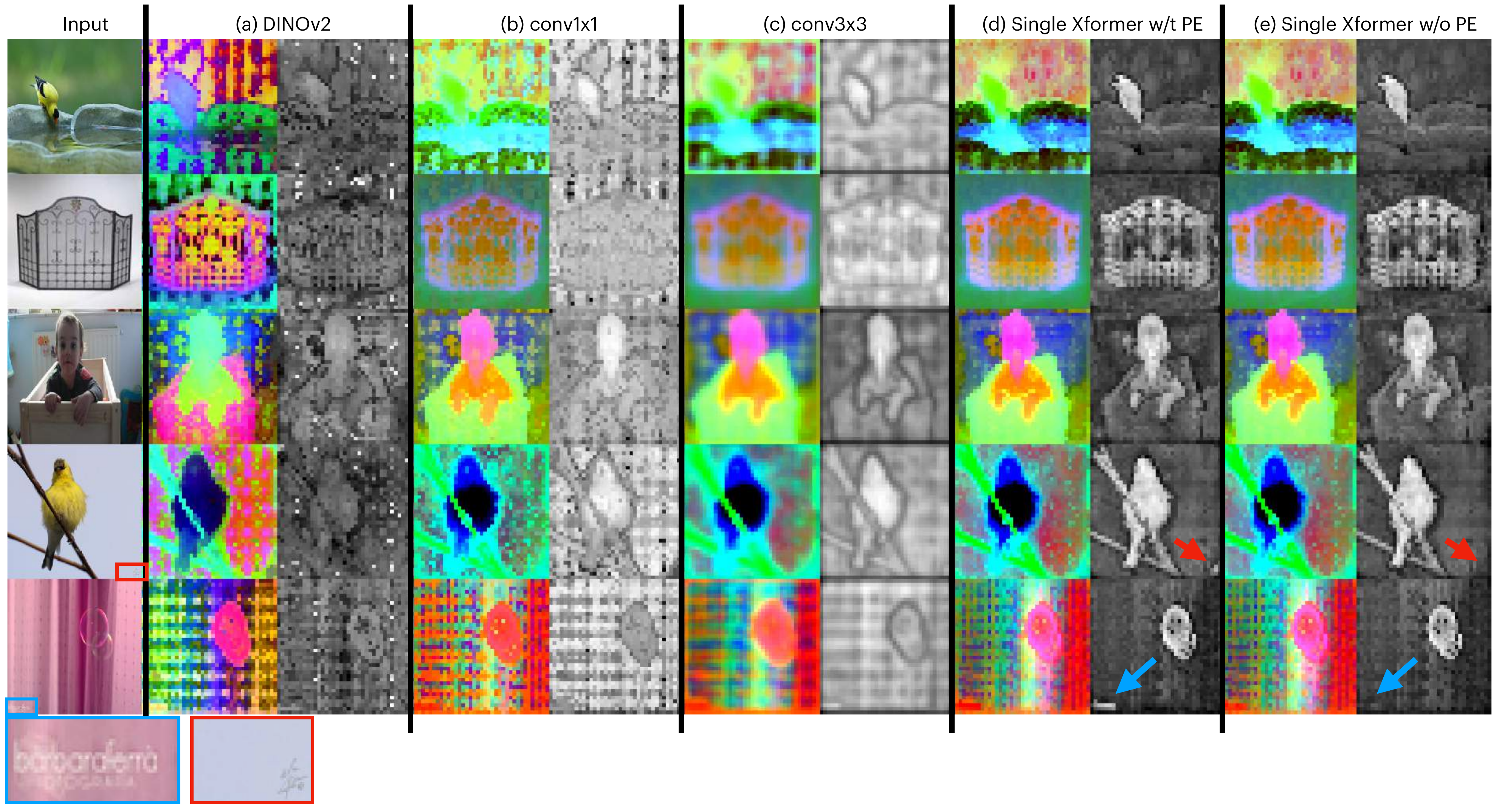}
    \caption{\textbf{Qualitative comparison of different denoiser architecture designs}. Convolution-based denoisers typically do not yield good performance (b, c). We empirically find that the denoiser with learnable new positional embeddings (PE) is sensitive to subtle details (see the blue and red rectangles and arrows). ``Xformer'': Transformer block. }
    \label{fig:arch_ablation}
\end{figure}
\paragraph{Optimization.} To train the denoiser, we optimize the loss function defined in \cref{eq:gen_loss} of the main text. Note that our approach does not necessitate re-training ViTs; instead, it only optimizes the smaller denoisier network, which constitutes only 8\% of the original model's size. The denoiser is trained for 10 epochs with a batch size of 64, using the \texttt{AdamW} optimizer with a learning rate of $2e-$4 and a cosine learning rate scheduler. The denoiser training typically takes about 2 hours on 8 GPUs.

\subsection{ViT Models}
\paragraph{Model identifiers.} We provide the \texttt{timm} model identifiers of the ViTs studied in this paper in \cref{tab:model_id}. For experiments with large input image sizes (\eg using the 512-sized images as input to a model trained with 224-image resolution), we always resize the position embeddings using bicubic interpolation to accommodate the increased size.
\begin{table*}
    \centering
    \small
    \caption{\textbf{\texttt{timm} model identifiers}}
    \tablestyle{3.5pt}{1.2}
    \footnotesize
    \begin{tabular}{l | y{185}}
        Model                            & Model identifier                                                \\
        \shline
        DINOv2 \cite{oquab2023dinov2}    & \texttt{vit\_base\_patch14\_dinov2.lvd142m}                     \\
        Register \cite{darcet2023vision} & \texttt{vit\_base\_patch14\_reg4\_dinov2.lvd142m}               \\
        DINO \cite{caron2021emerging}    & \texttt{vit\_base\_patch16\_224.dino}                           \\
        MAE \cite{he2022masked}          & \texttt{vit\_base\_patch16\_224.mae}                            \\
        EVA02 \cite{fang2023eva}         & \texttt{eva02\_base\_patch16\_clip\_224.merged2b}               \\
        CLIP \cite{radford2021learning}  & \texttt{vit\_base\_patch16\_clip\_384.laion2b\_ft\_in12k\_in1k} \\
        DeiT-III \cite{touvron2022deit}  & \texttt{deit3\_base\_patch16\_224.fb\_in1k}                     \\
    \end{tabular}
    \label{tab:model_id}
\end{table*}

\subsection{Correlation}
In the main text, we mention the correlation between artifacts and their positions in images without a detailed context, which we now provide. Our focus is on quantifying the correlation between different features and their positions within an image. To analyze this correlation, we employ the maximal information coefficient (MIC), a metric originally used for measuring the strength of linear or nonlinear associations between two scalar variables. To adapt MIC for our purpose, we compute the association between high-dimensional features $\mathbf{f}$ and their positions. We calculate this by taking the maximal MIC across all channels of $\mathbf{f}$ and averaging the MICs of the coordinates $x$ and $y$.
\begin{equation}
    \frac{\max_{c \in \mathcal{C}}{\text{MIC}(\mathbf{f}(x, :), x)} + \max_{c \in \mathcal{C}}{\text{MIC}(\mathbf{f}(:, y), y)}}{2},
\end{equation}
where $\mathbf{f}(x, :)$ denotes the feature vector on the $x$-coordinate, $\mathbf{f}(:, y)$ at the $y$-coordinate, and $\mathcal{C}$ is the channel size of $\mathbf{f}$. For hyperparameters of scalar MIC, we set $B=(H \times W)^{0.6}$:
\begin{equation}
    \text{MIC}(\mathbf{X}; \mathbf{Y}) = \max_{|\mathbf{X}| |\mathbf{Y}| < B} \frac{I[\mathbf{X}; \mathbf{Y}]}{\log_2{(\min{(|\mathbf{X}|, |\mathbf{Y}|}))}},
\end{equation}
where $I[\mathbf{X}; \mathbf{Y}]$ denotes the mutual information between two random variables $\mathbf{X}$ and $\mathbf{Y}$. We compute this metric from 100 randomly selected samples from the ImageNet dataset.

Our analysis includes a comparison of the MIC values for the decomposed noise map, the original noisy ViT features, and the denoised, artifact-free features. The results, presented in \cref{tab:correlation} of the main paper, reveal that the decomposed noise map exhibits the highest correlation with image positions. The noisy features, which are entangled with noise artifacts originating from the position embeddings, display the second highest positional correlation. In contrast, the noise-free features denoised by our method show the lowest correlation with positions, demonstrating the effectiveness of our decomposition approach in removing such artifacts.

\subsection{Feature Qualitative Results}

\begin{figure}[t]
    \centering
    \includegraphics[width=0.85\linewidth]{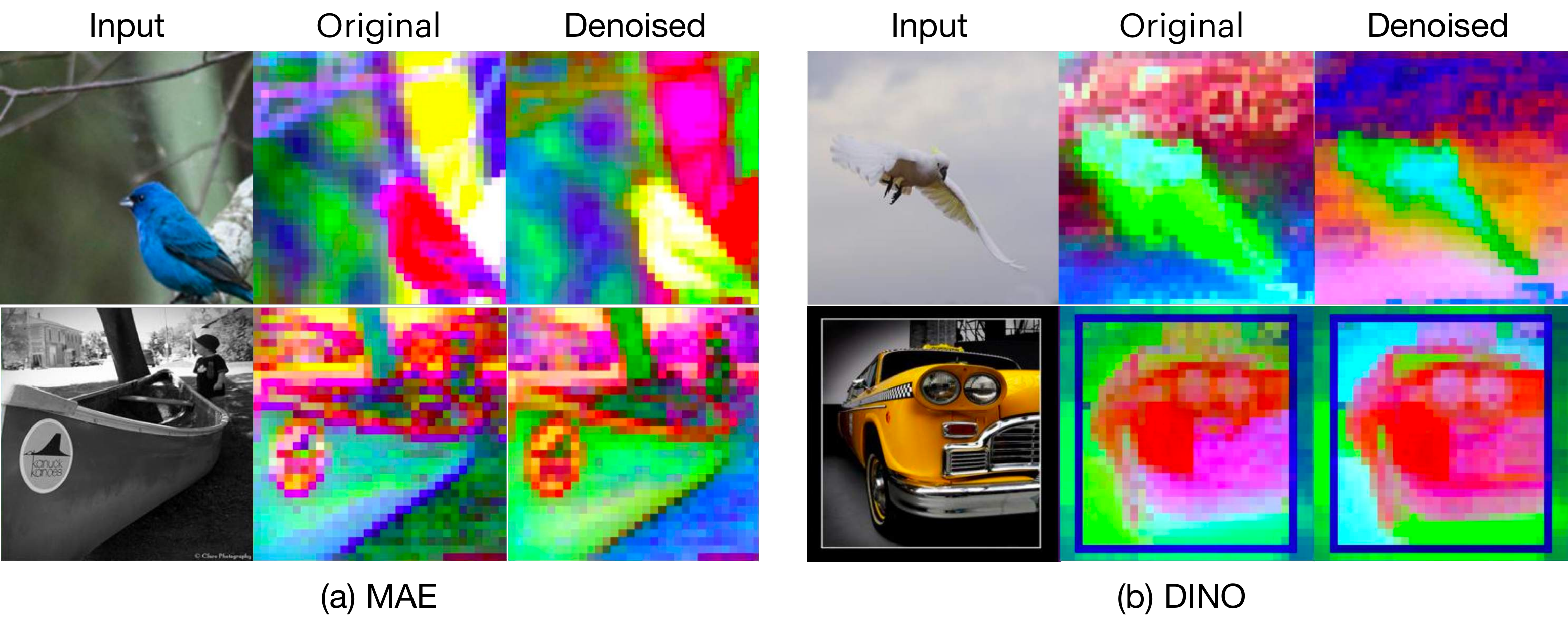}
    \caption{
    \textbf{Features from Weak Artifact Algorithms.}
    }
    \label{fig:weak-artif}

\end{figure}

\begin{figure}
    \centering
    \includegraphics[width=\linewidth]{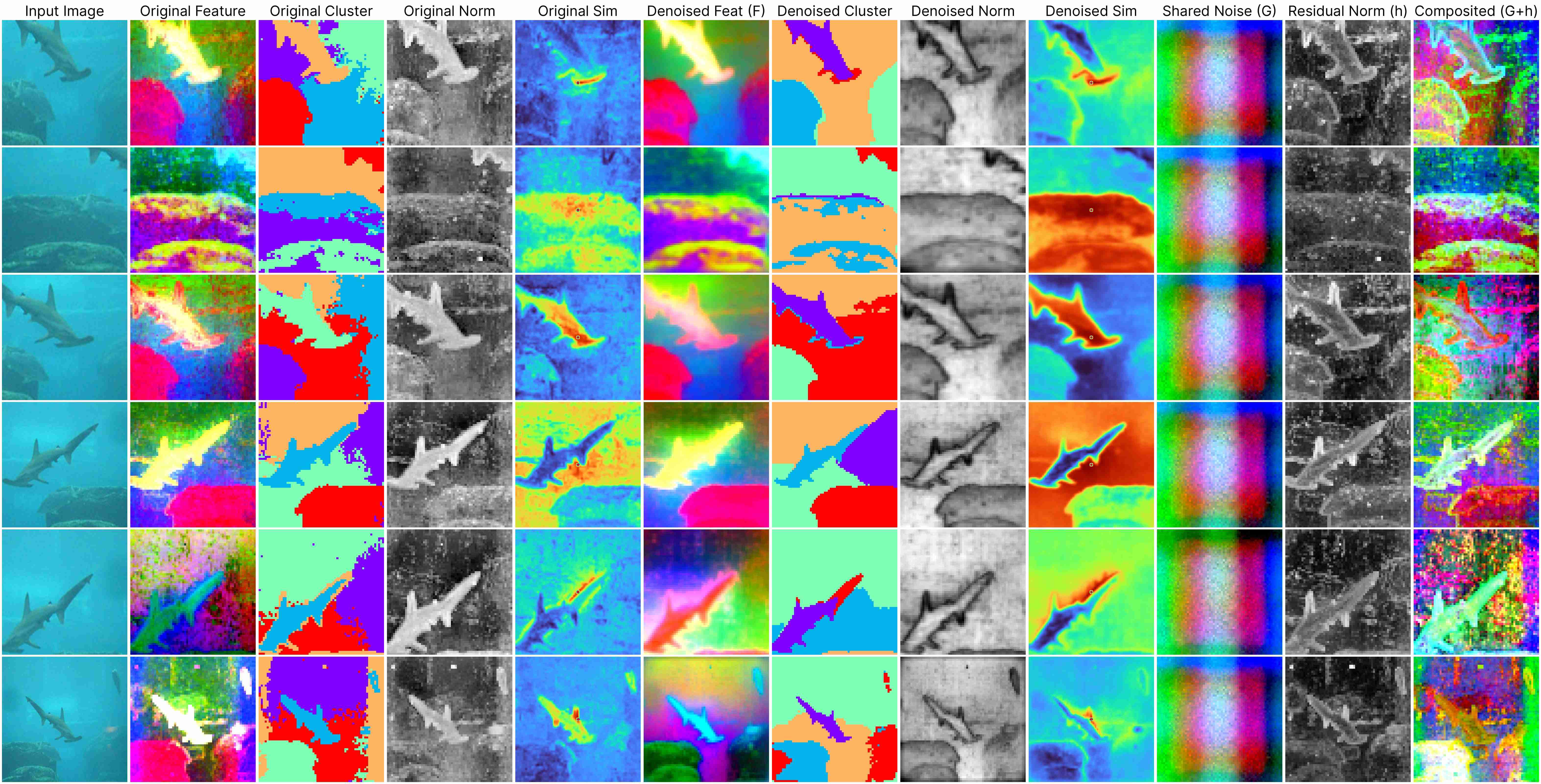}
    \caption{\textbf{Visualization of DINO \cite{caron2021emerging} per-image denoising}. We visualize all components of the per-image denoising stage. From left to right: In the first 5 columns we visualize the input image, the original noisy feature map from the model, the K-Means clusters on the original features, the L2 norm on the original features, and the similarity between the central red patch and other patches. In the next 4 columns we visualize the the denoised feature map using \ours{}, the denoised features' K-means clusters, the denoised features' L2 norms, and their similarity post-denoising. In the last 3 columns we visualize the decomposed shared noise term $\mathcal{G}$, the L2 norm of the predicted residual term $\vb{h}$, and the composite noise $(\mathcal{G} + \vb{h})$.}
    \label{fig:per_img_dino}
\end{figure}

\begin{figure}
    \centering
    \includegraphics[width=\linewidth]{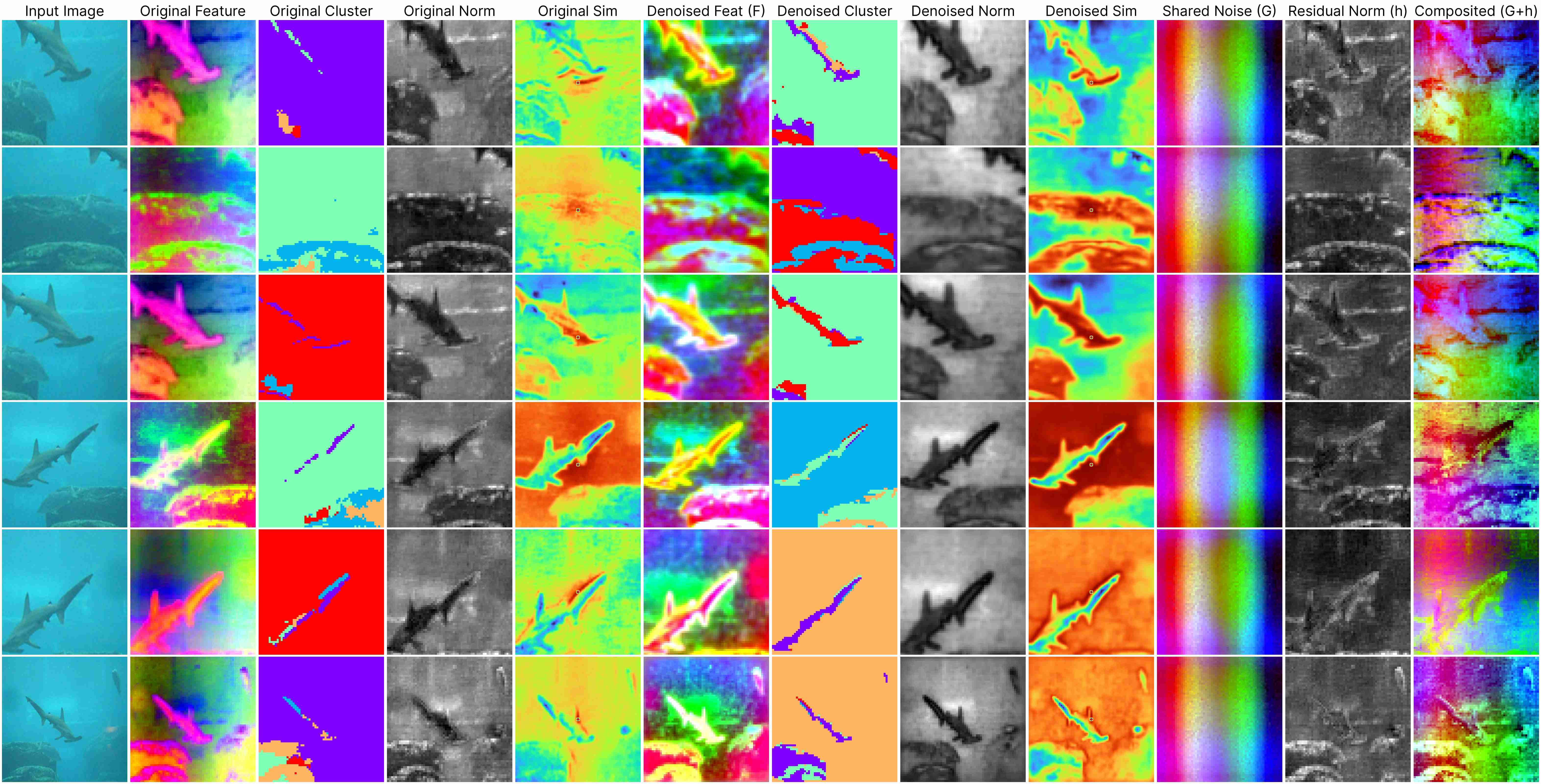}
    \caption{\textbf{Visualization of MAE \cite{he2022masked} per-image denoising}. We visualize all components of the per-image denoising stage. From left to right: In the first 5 columns we visualize the input image, the original noisy feature map from the model, the K-Means clusters on the original features, the L2 norm on the original features, and the similarity between the central red patch and other patches. In the next 4 columns we visualize the the denoised feature map using \ours{}, the denoised features' K-means clusters, the denoised features' L2 norms, and their similarity post-denoising. In the last 3 columns we visualize the decomposed shared noise term $\mathcal{G}$, the L2 norm of the predicted residual term $\vb{h}$, and the composite noise $(\mathcal{G} + \vb{h})$.}
    \label{fig:per_img_mae}
\end{figure}

\paragraph{Algorithms producing mild artifacts.}
We additionally visualize the features for algorithms with weak artifacts in \cref{fig:weak-artif}.
We empirically observe that ViTs trained using both MAE and DINO exhibit very few visible artifacts in their feature (center column). \cref{fig:per_img_dino,fig:per_img_mae} shows additional visualizations of the decomposed noise map and the learned residual terms of DINO and MAE, respectively. We note that decomposed noise maps from these two models typically manifest low-frequency patterns and the residual terms do not yield pronounced patterns.

\begin{figure}
    \centering
    \includegraphics[width=\linewidth]{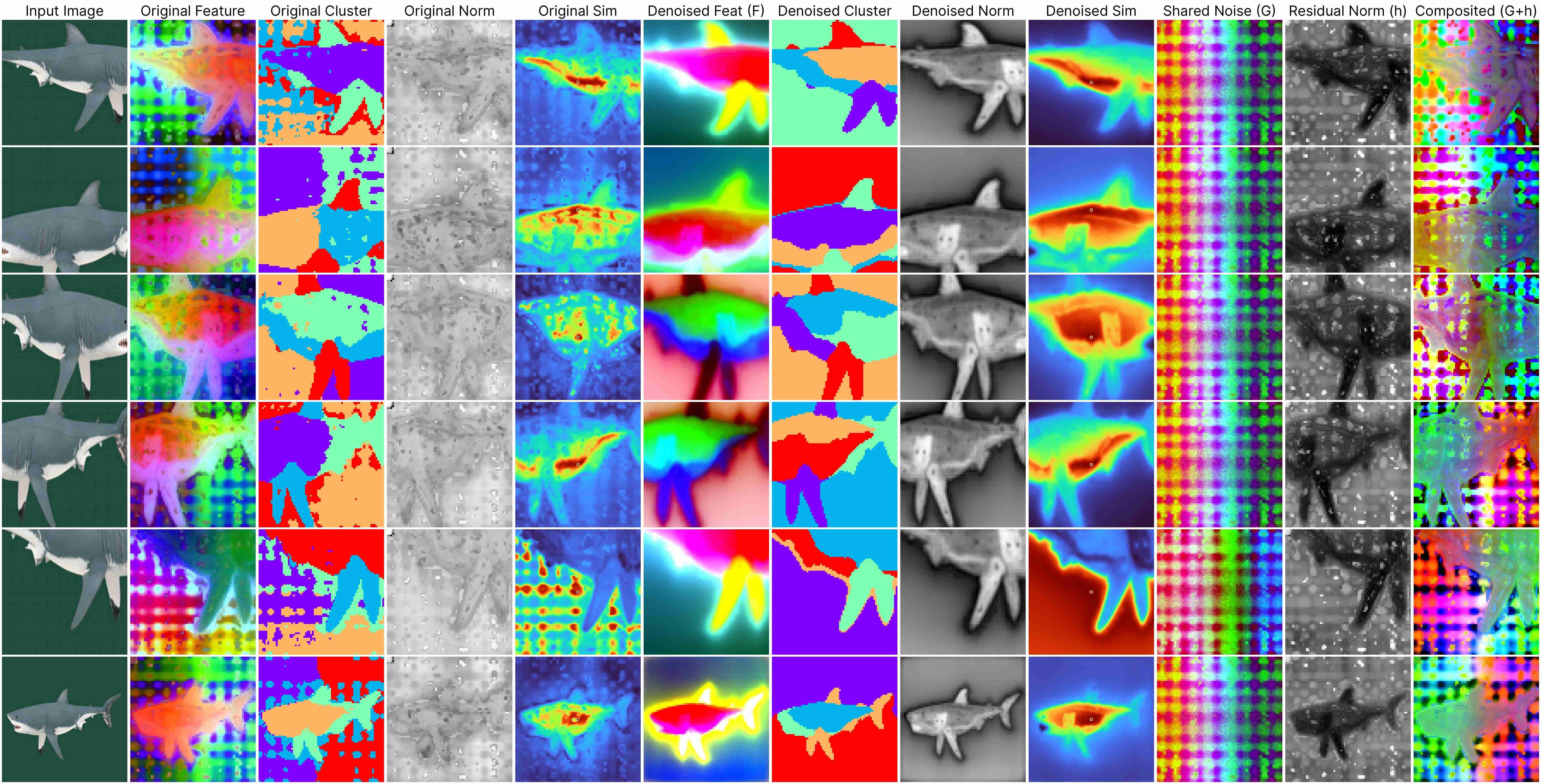}
    \caption{\textbf{Visualization of DINOv2 \cite{oquab2023dinov2} per-image denoising}. We visualize all components of the per-image denoising stage. From left to right: In the first 5 columns we visualize the input image, the original noisy feature map from the model, the K-Means clusters on the original features, the L2 norm on the original features, and the similarity between the central red patch and other patches. In the next 4 columns we visualize the the denoised feature map using \ours{}, the denoised features' K-means clusters, the denoised features' L2 norms, and their similarity post-denoising. In the last 3 columns we visualize the decomposed shared noise term $\mathcal{G}$, the L2 norm of the predicted residual term $\vb{h}$, and the composite noise $(\mathcal{G} + \vb{h})$.}
    \label{fig:per_img_dinov2}
\end{figure}

\begin{figure}
    \centering
    \includegraphics[width=\linewidth]{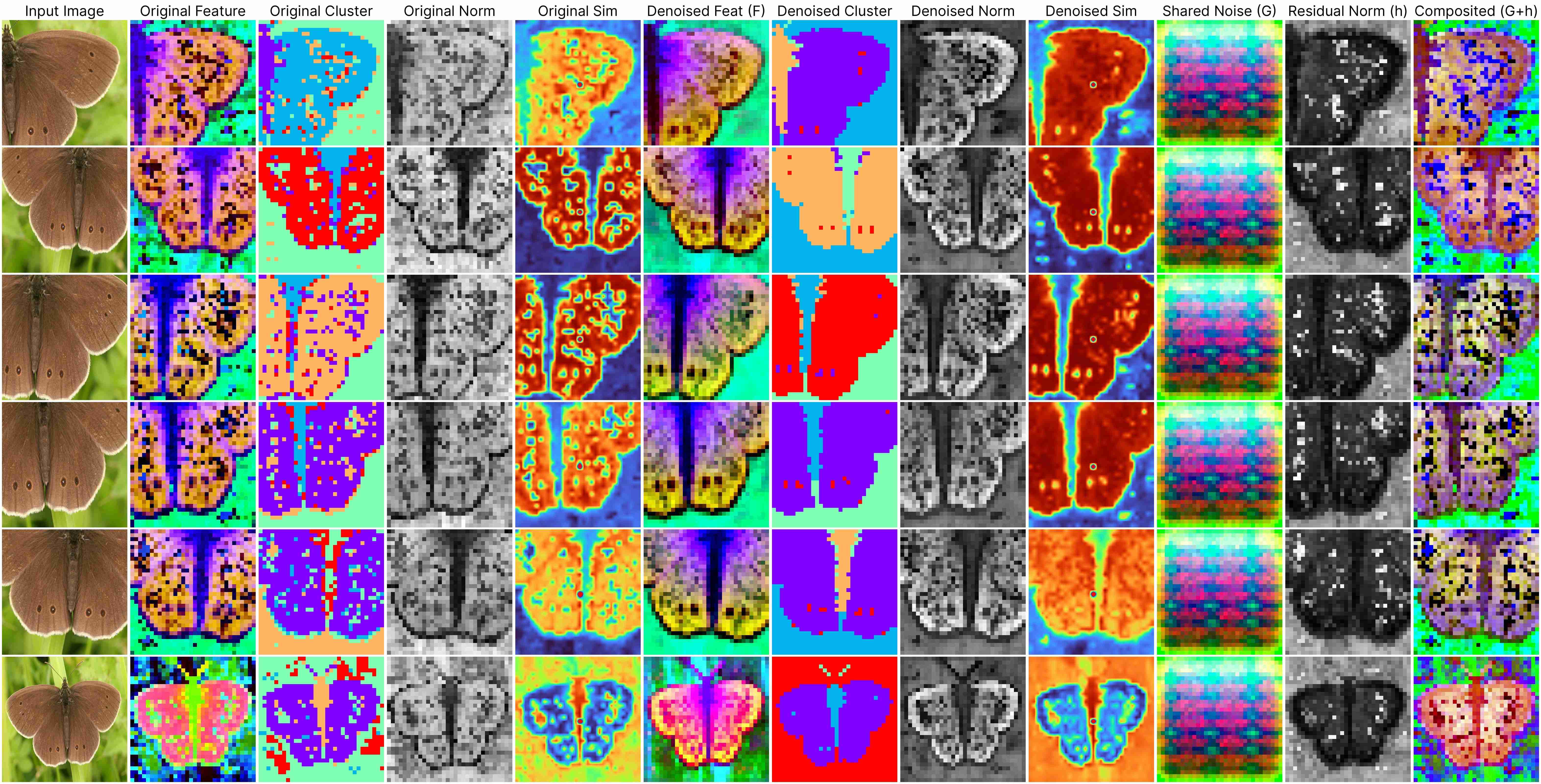}
    \caption{\textbf{Visualization of CLIP \cite{radford2021learning} per-image denoising}. We visualize all components of the per-image denoising stage. From left to right: In the first 5 columns we visualize the input image, the original noisy feature map from the model, the K-Means clusters on the original features, the L2 norm on the original features, and the similarity between the central red patch and other patches. In the next 4 columns we visualize the the denoised feature map using \ours{}, the denoised features' K-means clusters, the denoised features' L2 norms, and their similarity post-denoising. In the last 3 columns we visualize the decomposed shared noise term $\mathcal{G}$, the L2 norm of the predicted residual term $\vb{h}$, and the composite noise $(\mathcal{G} + \vb{h})$.}
    \label{fig:per_img_clip}
\end{figure}

\begin{figure}
    \centering
    \includegraphics[width=\linewidth]{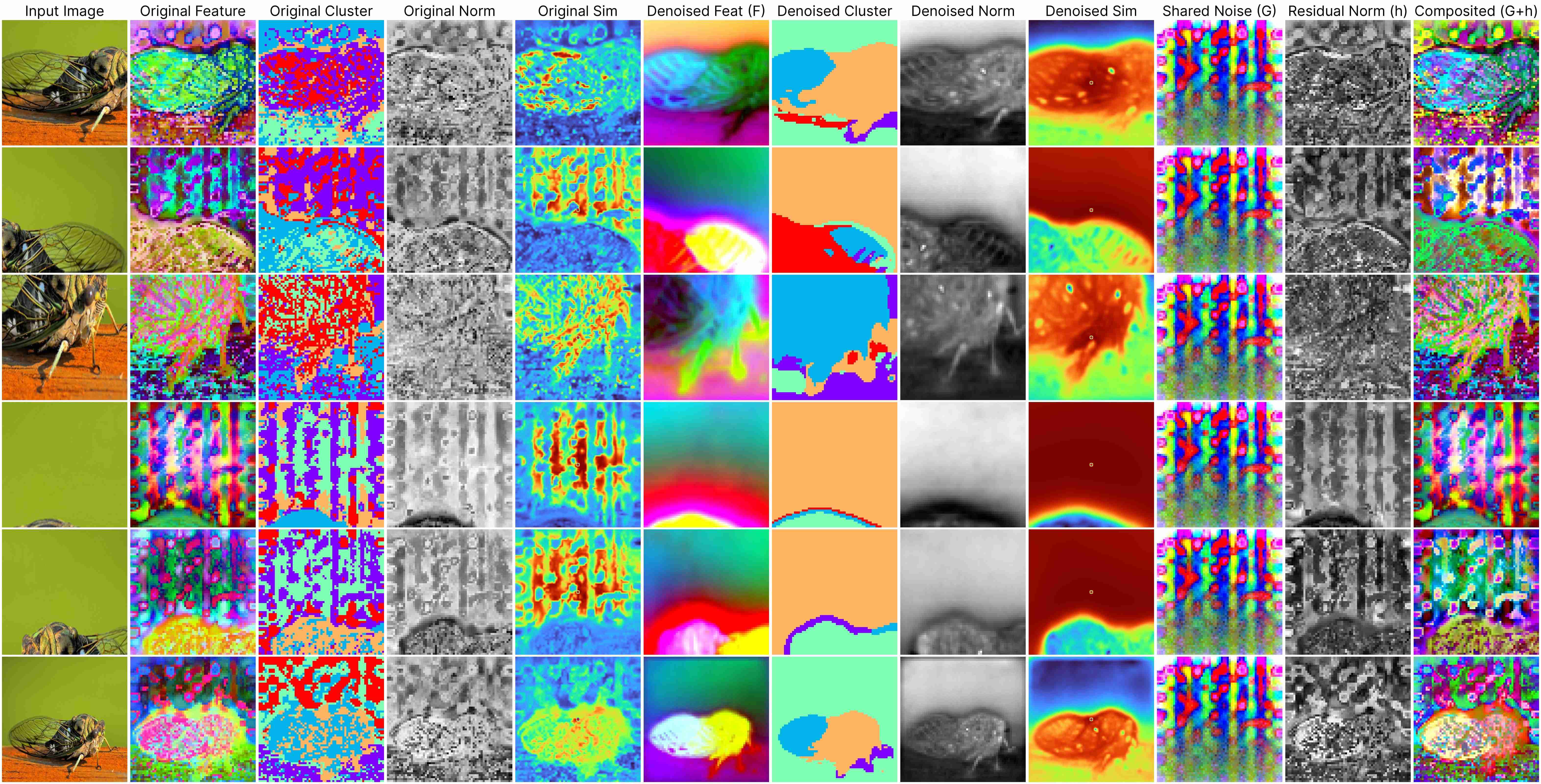}
    \caption{\textbf{Visualization of EVA02 \cite{fang2023eva} per-image denoising}. We visualize all components of the per-image denoising stage. From left to right: In the first 5 columns we visualize the input image, the original noisy feature map from the model, the K-Means clusters on the original features, the L2 norm on the original features, and the similarity between the central red patch and other patches. In the next 4 columns we visualize the the denoised feature map using \ours{}, the denoised features' K-means clusters, the denoised features' L2 norms, and their similarity post-denoising. In the last 3 columns we visualize the decomposed shared noise term $\mathcal{G}$, the L2 norm of the predicted residual term $\vb{h}$, and the composite noise $(\mathcal{G} + \vb{h})$.}
    \label{fig:per_img_eva}
\end{figure}

\begin{figure}
    \centering
    \includegraphics[width=\linewidth]{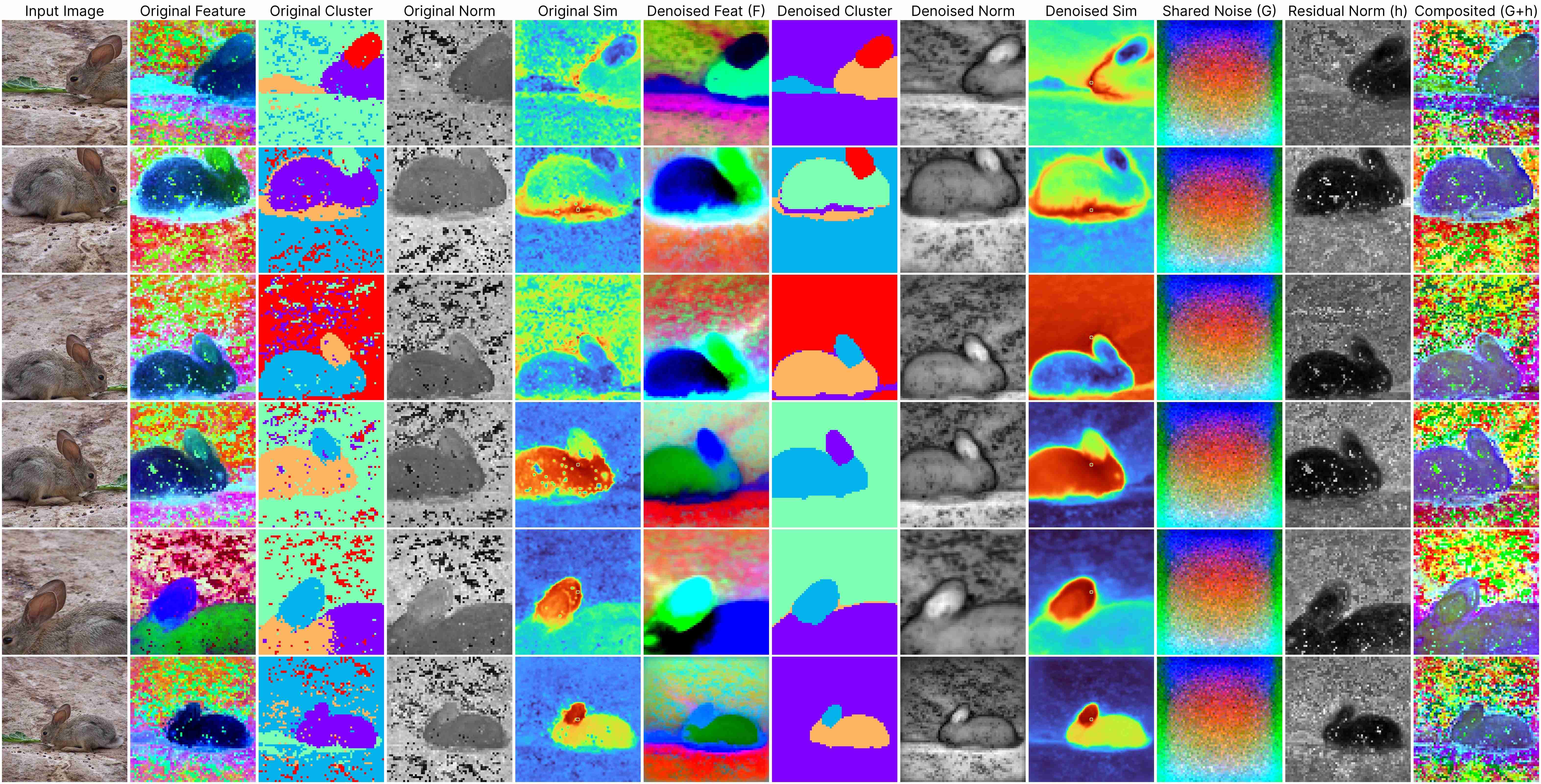}
    \caption{\textbf{Visualization of DeiT-III \cite{touvron2022deit} per-image denoising}. We visualize all components of the per-image denoising stage. From left to right: In the first 5 columns we visualize the input image, the original noisy feature map from the model, the K-Means clusters on the original features, the L2 norm on the original features, and the similarity between the central red patch and other patches. In the next 4 columns we visualize the the denoised feature map using \ours{}, the denoised features' K-means clusters, the denoised features' L2 norms, and their similarity post-denoising. In the last 3 columns we visualize the decomposed shared noise term $\mathcal{G}$, the L2 norm of the predicted residual term $\vb{h}$, and the composite noise $(\mathcal{G} + \vb{h})$.}
    \label{fig:per_img_deit}
\end{figure}

\begin{figure}
    \centering
    \includegraphics[width=\linewidth]{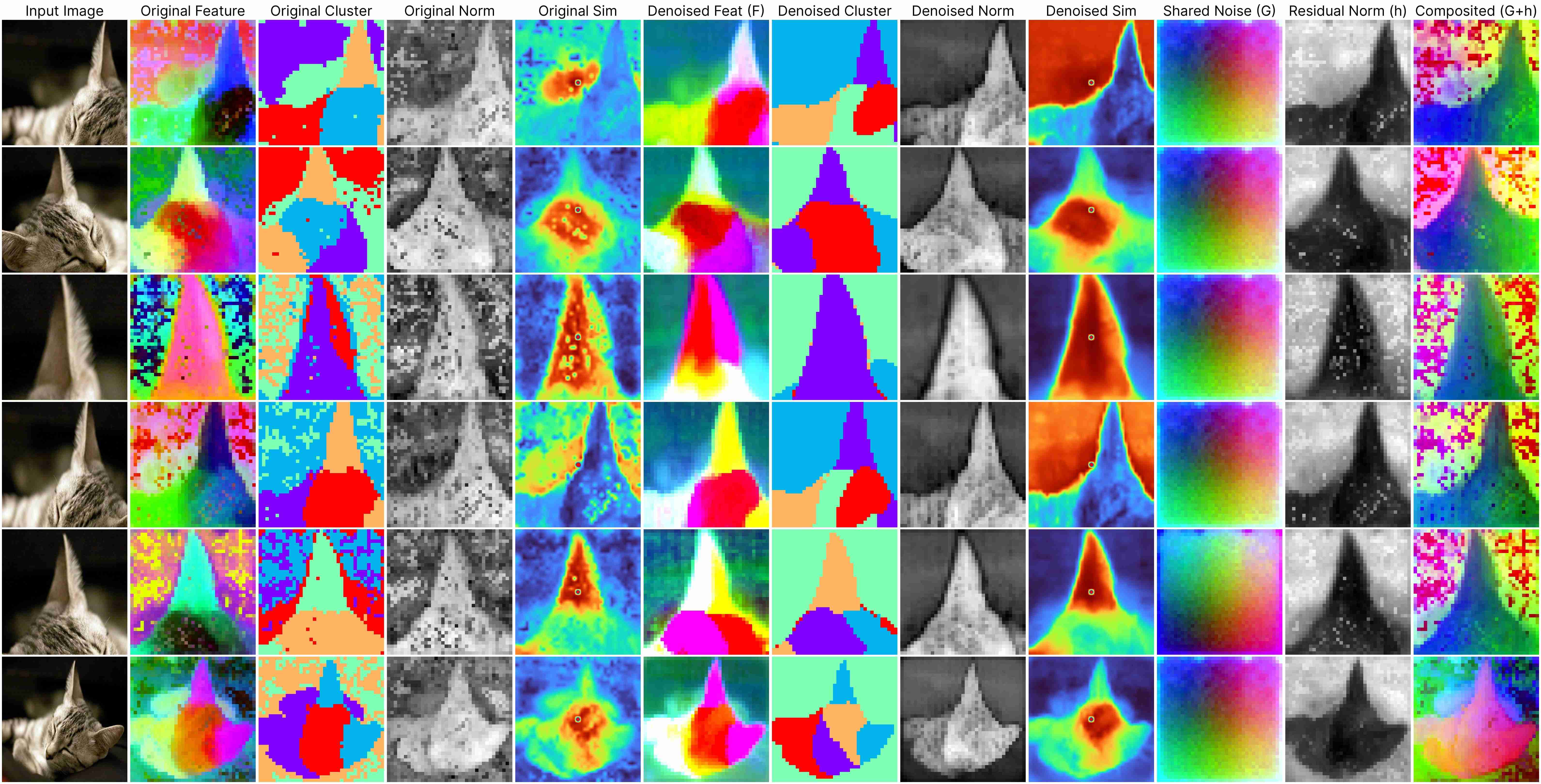}
    \caption{\textbf{Visualization of DINOv2 with Registers \cite{darcet2023vision} per-image denoising}. We visualize all components of the per-image denoising stage. From left to right: In the first 5 columns we visualize the input image, the original noisy feature map from the model, the K-Means clusters on the original features, the L2 norm on the original features, and the similarity between the central red patch and other patches. In the next 4 columns we visualize the the denoised feature map using \ours{}, the denoised features' K-means clusters, the denoised features' L2 norms, and their similarity post-denoising. In the last 3 columns we visualize the decomposed shared noise term $\mathcal{G}$, the L2 norm of the predicted residual term $\vb{h}$, and the composite noise $(\mathcal{G} + \vb{h})$.}
    \label{fig:per_img_register}
\end{figure}

\paragraph{Additional visualizations.}
Additional visualizations of the feature maps at all layers of ViT models are shown in \cref{fig:strong_multilayer}. Observe that the artifact is present in almost all layers of the models. See \cref{fig:per_img_dinov2,fig:per_img_clip,fig:per_img_deit,fig:per_img_eva,fig:per_img_mae,fig:per_img_dino,fig:per_img_register} for more visualizations.

\section{Evaluation Protocols}
\label{sec:evaluation_protos}

We introduce our evaluation protocols here, mostly following \cite{oquab2023dinov2}. We have released our code, checkpoints, and logs for reproducibility at \url{https://jiawei-yang.github.io/DenoisingViT/}.

\paragraph{Semantic Segmentation.} We use a linear evaluation setting. In detail, we extract the final feature maps from the frozen backbone and pass them through the denoisers if there are any. Following this, feature maps are resized back to their original resolution. Then, a single learnable linear layer is trained to predict the semantic segmentation from these resized feature maps. The training and testing image resolutions are $518\times 518$, following \cite{oquab2023dinov2}. We train this linear head for $40,000$ iterations for both VOC and ADE20k datasets. We report the mean intersection over union (mIoU) metric for all experiments.

\paragraph{Depth estimation.} We extract the final feature maps from the frozen backbone and pass them through the denoisers, if applicable. Then, we follow \cite{oquab2023dinov2} to append the \texttt{cls} token to every patch token to enrich feature representations. We bilinearly upsample these features by a factor of 4 and train a linear layer using classification loss to divide the prediction into 256 uniformly distributed bins. Unlike \cite{oquab2023dinov2}, we slightly decrease the learning rate from 1e-4 to 5e-3, as we find that this modification improves most of the methods, including baselines, in our early experiments.
We report our results on the commonly used metrics: AbsRel (absolute relative error $\abs{d^{*} - d}/d$) and $\delta_1$ (percentage of pixels where $\max(d^{*}/d,~ d/d^{*}) < 1.25$).

\paragraph{Object detection.} To evaluate the object detection task, we utilize the ViTDet detector \cite{li2022exploring} to infer object bounding boxes based on feature maps extracted either from original ViTs or denoisers. The detection framework is FasterRCNN \cite{ren2015faster}. The input image resolution for training and testing is $518\times 518$, the same as the semantic segmentation task. We train all models for 24k iterations, where we decay the learning rate at steps 20k and 22k.

Our initial attempts at directly learning an object detection head from the denoised features did not achieve superior performance. This led us to speculate that the omission of relative positional information, which was largely removed during denoising, might be important for accurately predicting the relative box coordinates of objects within the full image context. This requirement is almost \textit{unique} to the bounding box prediction task. To counteract this, we re-add fixed sinusoidal positional embeddings into the feature maps produced by the denoisers. This adjustment, adding no additional learnable parameters, is found to enhance the detection performance. We believe that the disentanglement between positional features and semantic features would be an interesting direction to study. We also apply this method to the baseline models, and the results are shown in \cref{tab:obj_det_more}. We see that adding this step to the baselines does not yield consistent performance gains. Consequently, we apply this step only to our denoisers.

\begin{table}[t!]
    \centering
    \caption{\textbf{Object detection with frozen features.} We report the mAP metric on the VOC object detection benchmark. ``fixed PE'': fixed sinusoidal positional embeddings.}
    \tablestyle{3.5pt}{1.2}
    \begin{tabular}{l | y{35} y{40} y{35} y{35} y{35} y{35} y{35}}
        Method                                  & DINOv2                                     & DINOv2-reg                                 & DeiT-III                                   & CLIP                                       & DINO                                       & EVA02                                      & Avg                                        \\
        \shline  baseline                       & 81.4                                       & 80.9                                       & 80.9                                       & 75.8                                       & 76.4                                       & 79.4                                       & 79.2                                       \\
        \hspace{.5em}+fixed PE                  & {81.5} \increase{(+0.1)}                   & {81.2} \increase{(+0.3)}                   & 80.9                                       & {75.7} \decrease{(-0.1)}                   & {75.8} \decrease{(-0.6)}                   & {78.7} \decrease{(-0.7)}                   & 79.0 \decrease{(-0.2)}                     \\
        \baseline{\hspace{.5em}+\textbf{\ours}} & \baseline{\textbf{81.9} \increase{(+0.5)}} & \baseline{\textbf{81.4} \increase{(+0.5)}} & \baseline{\textbf{81.7} \increase{(+0.9)}} & \baseline{\textbf{77.0} \increase{(+1.2)}} & \baseline{\textbf{77.1} \increase{(+0.7)}} & \baseline{\textbf{80.2} \increase{(+0.8)}} & \baseline{\textbf{79.9} \increase{(+0.7)}} \\
    \end{tabular}
    \label{tab:obj_det_more}
\end{table}

\paragraph{Object discovery.} We use LOST \cite{simeoni2021localizing} to evaluate the object discovery performance.
LOST leverages the activation features of a pre-trained ViT for automated object discovery. Specifically, it uses the components of the last attention layer for computing the similarities between the different patches to discover and identify the object connected components.
To use LOST, one has to manually sweep between query, key, value, or other intermediate model outputs as the indictor of objects' prominence. Through our qualitative analysis, we find that the feature norm is a good candidate to indict object prominence (See \cref{fig:emerg_obj_dis}). We report our results on the \texttt{CorLoc} metric (percentage of predicted box with an IoU greater than 0.5 with one of the labeled object bounding boxes) as in \cite{simeoni2021localizing,darcet2023vision}.

\paragraph{Classification.} Although the global-level classification task is beyond the scope of our approach, our \ours demonstrates improved performance over its baselines through the use of an attentive probe protocol. Following the methodology described in AIM \cite{el2024scalable}, we conduct an ``attentive probe'' on both the original and denoised patch tokens, omitting the \texttt{CLS} token, which our approach does not process during training. This probe employs attention mechanisms to maximize the extraction of information from each patch token. The backbones and the denoisers are frozen during our evaluation, and we train the attentive layer for 10 epochs. The results, presented in \cref{tab:classification}, suggest that DVT can potentially improve over its baselines, even though the denoising objective is orthogonal to classification. We believe integrating the \texttt{CLS} token into the denoising process represents a promising avenue for future research to enhance classification performance.

Additionally, we underscore the versatility of the denoiser in our \ours as a \textit{plug-in-and-play} module, which can be optionally activated or deactivated to support various functionalities without compromising \textit{any} properties of the original models. In essence, by leveraging the original class tokens before the denoiser, one can always recover the original models' classification performance.

\begin{table}[t!]
    \centering
    \small
    \caption{\textbf{ImageNet Classification Accuracy using Attentive Probing.}}
    \tablestyle{3.5pt}{1.2}
    \begin{tabular}{l | y{40} y{45} y{40} y{40} y{40} y{40}}
        Method                                       & DINOv2 & DINOv2-reg & DeiT-III & CLIP   & DINO   & EVA02  \\
        \shline
        baseline                                     & 81.2\% & 81.6\%     & 81.6\%   & 83.4\% & 77.0\% & 80.3\% \\
        \baseline{\hspace{.5em}+\ours}               &
        \baseline{\textbf{81.8}\% \increase{(+0.6)}} &
        \baseline{\textbf{81.8}\% \increase{(+0.2)}} &
        \baseline{\textbf{82.1}\% \increase{(+0.5)}} &
        \baseline{\textbf{83.5}\% \increase{(+0.1)}} &
        \baseline{77.0\%}                            &
        \baseline{\textbf{80.5}\% \increase{(+0.2)}}
    \end{tabular}
    \label{tab:classification}
\end{table}

\section{Further Discussion into ViT Understanding}
\label{sec:understanding}

\paragraph{Different positional embeddings.} The models studied in this paper cover three major types of position embeddings (PEs) --- fixed sinusoidal PE (\eg, MAE \cite{he2022masked}), learnable additive PE (\eg, DINO \cite{caron2021emerging}, DINOv2 \cite{oquab2023dinov2}, CLIP \cite{radford2021learning}, DeiT-III \cite{touvron2022deit}), and learnable Rotary PE (\eg EVA02 \cite{fang2023eva}). Intriguingly, our observations reveal that, regardless of the type of PE employed, artifacts are present in all the studied ViTs, though with varying extents. The emergence of artifacts seems to be a common characteristic across different PE types. Although the fundamental underlying reason behind this property remains unclear, our work identifies this issue and proposes a denoising method to rectify these artifacts.

\paragraph{Alternative approaches for position embeddings.} A key component of our hypothesis as to why artifacts exist in ViT features is the use of positional embeddings. Currently, all ViTs leverage either fixed \cite{he2022masked} or learned \cite{Caron_2021, oquab2023dinov2, touvron2022deit} positional embeddings that are \textit{added to the input tokens} of the Transformer model.
Alternatively, Rotary Positional Embeddings \cite{su2023roformer}, which were originally proposed in the language domain for better sequence length scaling, does not directly add anything to the input tokens. Instead, this method encodes the absolute position with a rotation matrix and crucially incorporates the explicit relative position dependency in the computation of the attention values.
Although EVA02 \cite{fang2023eva} does leverage this kind of positional embedding, the training process involves distilling from the already-noisy features of CLIP. Indeed, the noisy artifacts of the EVA02 model resemble those of CLIP models, especially in the later layers (\cref{fig:strong_multilayer}).
Thus, while the positional embedding selection is promising, more research should be done towards ViTs that leverage these Rotary PE for artifact reduction.
Similarly, the positional embedding used in the T5 language model \cite{raffel2023exploring} does not add a positional embedding directly to the input; instead, it learns a bias that is added to the key-query dot product in the self-attention step and does not include explicit position information in the self-attention value vectors. ALiBi \cite{press2022train}, used in many large language models (LLM), also does not do so, and instead adds a static bias to the query-key dot product. These methods eliminate the input-independent portion of the final output feature while retaining the benefits of the position embedding. For future work, we suggest further exploration into adapting other such positional embedding paradigms specifically for the image domain.


\section{Discussion on Limitations}
\label{sec:limitation}
\paragraph{Limitations.} Our approach faces some practical and theoretical challenges. On the practical front, although our method leverages parallel computing to amortize the denoising process, the time required to denoise a single image, such as one with a resolution of $518 \times 518$, remains high --- approximately 100 seconds. This duration may be impractical for commercial or personal users with limited access to parallel computing resources, despite the fact that we can finish denoising 10k samples within hours. Additionally, our generalizable denoisesr, trained on the last layer features of pretrained ViTs, does not remove noise in intermediate outputs. Users requiring denoised features from multiple layers might need to train distinct denoisers for different layers. From the theoretical perspective, the reasons behind the presence of these artifacts remain unclear. Integrating insights from Registers \cite{darcet2023vision} with our findings could yield a more comprehensive understanding of these phenomena.

\paragraph{Broader Impact.} Our work serves as one of the initial studies to understand the position-based artifacts present in the features of ViT models. We identify and propose methods to mitigate these artifacts, yet the root causes and characteristics of these artifacts are not fully understood. The severity of artifacts varies with the training algorithms; for instance, DINOv2 exhibits more pronounced artifacts compared to MAE, which shows subtler discrepancies.
Thus, one direction of exploration is to investigate the training paradigm that includes supervision ---\ie local \vs global --- as well as the loss-induced parameter landscape ---\ie sharp \vs smooth Hessians.
Furthermore, a better architectural design---\eg new positional embeddings---may diminish the severity of the feature artifacts. In this work, we do not explore modifying the ViT's design; however, more study into its positional embeddings and the effect on downstream features should prove interesting.
Ultimately, we believe our findings are intriguing to the community and more research is needed to better understand this fundamental problem.

\end{document}